\documentclass[]{fairmeta}

\usepackage[table]{xcolor}
\usepackage{pifont}
\usepackage{multirow}
\usepackage{hyperref}
\usepackage{url}
\usepackage{graphicx}
\usepackage{listings}
\usepackage{longtable}
\usepackage{arydshln}
\usepackage{caption}  %
\usepackage{subcaption} %
\usepackage{xspace}
\usepackage{adjustbox}
\usepackage{wrapfig}
\usepackage{algorithm}
\usepackage{algpseudocode}
\usepackage{enumitem}
\usepackage{booktabs}
\usepackage{array}
\usepackage{ulem}
\usepackage{tocloft} %
\usepackage{tabularx}

\newcommand{\dataset}{\textsc{FronTalk}}
\newcommand{\method}{\textsc{AceCoder}}
\newcommand{\red}[1]{{\color{red}#1}}

\definecolor{LinkBlue}{RGB}{0,0,238}
\definecolor{darkgold}{rgb}{0.72, 0.53, 0.04}

\definecolor{mygreen}{HTML}{00A64F}

\title{\dataset{}: Benchmarking Front-End Development as Conversational Code Generation with Multi-Modal Feedback}

\author[1,2,\ddagger]{Xueqing Wu}
\author[2]{Zihan Xue}
\author[1]{Da Yin}
\author[3,\ddagger]{Shuyan Zhou}
\author[2]{Kai-Wei Chang}
\author[2]{Nanyun Peng}
\author[1]{Yeming Wen}

\affiliation[1]{Meta Superintelligence Labs}
\affiliation[2]{University of California, Los Angeles}
\affiliation[3]{Duke University}

\contribution[\ddagger]{Work done at MSL}

\abstract{We present \textbf{\dataset{}}, a benchmark for front-end code generation that pioneers the study of a unique interaction dynamic: \textbf{conversational code generation with multi-modal feedback}. In front-end development, visual artifacts such as sketches, mockups and annotated screenshots are essential for conveying design intent, yet their role in multi-turn code generation remains largely unexplored. To address this gap, we focus on the front-end development task and curate \dataset{}, a collection of 100 multi-turn dialogues derived from real-world websites across diverse domains such as news, finance, and art. Each turn features both a textual instruction and an equivalent visual instruction, each representing the same user intent. To comprehensively evaluate model performance, we propose a novel \textit{agent-based evaluation framework} leveraging a web agent to simulate users and explore the website, and thus measuring both functional correctness and user experience. Evaluation of 20 models reveals two key challenges that are under-explored systematically in the literature: (1) a significant \textit{forgetting issue} where models overwrite previously implemented features, resulting in task failures, and (2) a persistent challenge in \textit{interpreting visual feedback}, especially for open-source vision-language models (VLMs). We propose a strong baseline to tackle the forgetting issue with \method{}, a method that critiques the implementation of every past instruction using an autonomous web agent. This approach significantly reduces forgetting to \textbf{nearly zero} and improves the performance by up to \textbf{9.3\%} (56.0\%$\rightarrow$65.3\%). Overall, we aim to provide a solid foundation for future research in front-end development and the general interaction dynamics of multi-turn, multi-modal code generation. Code and data are released at~~\url{https://github.com/shirley-wu/frontalk}
}

\date{\today}
\correspondence{Xueqing Wu at \email{xueqing.wu@cs.ucla.edu}}

\metadata[Code]{\url{https://github.com/shirley-wu/frontalk}}

\begin{document}

\maketitle

\begin{figure}[!h]
    \centering
    \includegraphics[width=\linewidth]{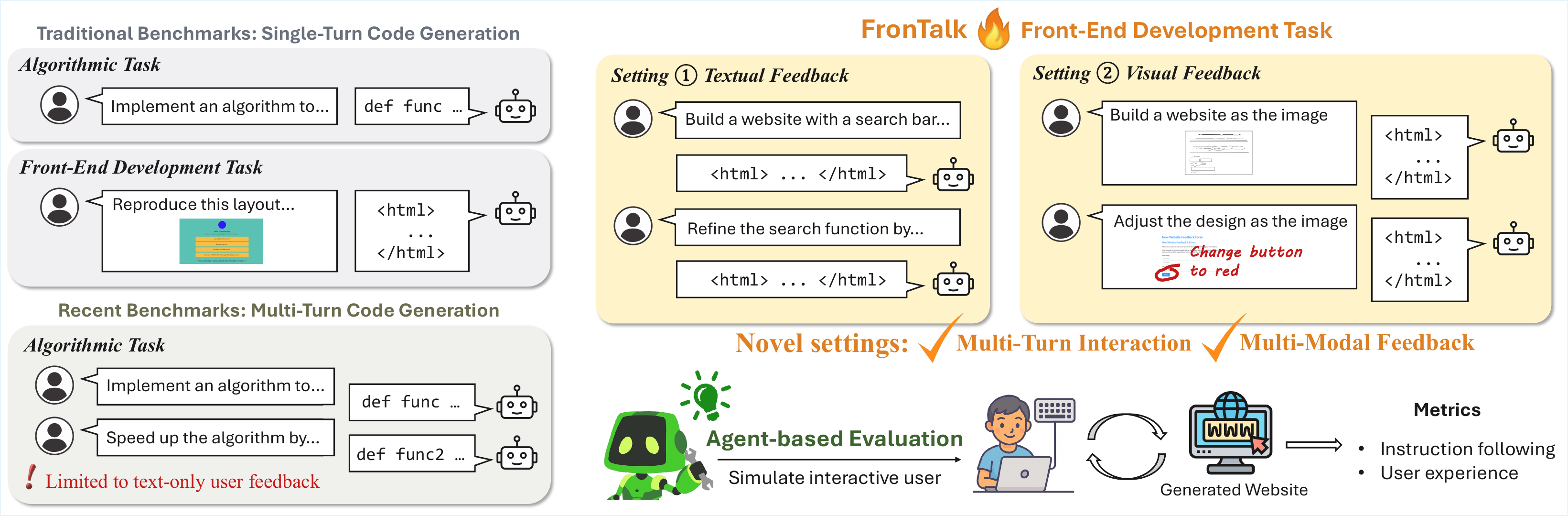}
    \caption{Focusing on the front-end development task, \textbf{\dataset{}} benchmark explores a novel setting with multi-turn interactions and multi-modal user feedback.}
    \label{fig:teaser}
\end{figure}

\section{Introduction}
\label{sec:intro}

The success of Large Language Models (LLMs) in code generation has popularized the application of AI-assisted coding involving \textbf{multi-turn interaction}, as seen in trends like ``vibe coding'' and a surge of commercial coding assistants. Reflecting this trend, recent benchmarking efforts have moved beyond single-turn code generation \citep{austin2021program,chen2021evaluating} to better measure how effectively LLMs respond to feedback across multiple turns \citep{wangmint,laban2025llms,hanconvcodeworld}. However, these benchmarks predominantly focus on algorithmic or API-driven tasks, where multi-modal feedback is less central, and thus limit their evaluation to text-only interactions. This results in under-exploration of domains such as data visualization, application design, and front-end development where \textbf{multi-modal feedback} is a critical communication channel. For example, in front-end development, sketching a UI layout or annotating a screenshot is a natural and efficient method for communicating design changes. Therefore, this unique dynamic of multi-turn, multi-modal coding remains a significantly underexplored challenge.

To advance this frontier, we introduce \dataset{}, a benchmark for front-end development that integrates multi-turn code generation with multi-modal feedback. Our benchmark begins with a diverse and realistic collection of \textit{user intents}, derived from real-world websites in the C4 dataset \citep{raffel2020exploring} and spanning diverse domains such as e-commerce platforms, financial service sites, and digital art portfolios. To capture realistic conversational dynamics, we employ an LLM-based \textit{user simulator} that generates context-aware instructions conditioned on prior dialogue. Crucially, to enable our simulated users to generate visual feedback, we equip them with a suite of \textit{drawing tools} to communicate intents through sketching and annotating. The resulted \dataset{} dataset comprises 1,000 conversational turns across 100 dialogues, paired with 3,676 manually refined test cases, offering a robust foundation for evaluating multi-modal, multi-turn coding systems. %

For evaluation, we propose an \textbf{agent-based evaluation framework} to evaluate both \textit{instruction following} and \textit{user experience}. For instruction following, we measure \textit{pass rate} against human-annotated test cases using an automated web agent to interact with the website and verify task completion. This agent is uniquely equipped with \textit{image manipulation tools} (e.g., cropping, image comparison), enhancing model's visual perception capabilities and improving alignment by 1.7\% with human annotators on design tasks.
For the nuanced dimension of user experience, we employ a \textit{pairwise comparison} protocol. In this setup, first-time users simulated by LLMs interact with each website to learn the interface and attempt self-proposed tasks. A secondary LLM then compares the resulting trajectories and judges which interface is more \textit{usable}, with criteria grounded in learning efficiency and satisfaction in task completion. Our evaluation protocol achieves a Cohen's Kappa of 0.63 for pass rate and 0.67 for usability, indicating significant alignment.

We evaluated a range of open- and closed-source models on our \dataset{} benchmark, including seven text-only LLMs evaluated on textual feedback and seven VLMs evaluated on both textual and visual feedback. Our experimental results reveal several key findings:
\begin{enumerate}[leftmargin=*, itemsep=0pt, topsep=0pt]
    \item \textbf{Open-source vs. proprietary gap}: As shown in Table \ref{tab:main}, proprietary models maintain a substantial lead over open-source models. This disparity is particularly pronounced with visual feedback, where the performance gap is 24.1\%, compared to the 12.5\% gap with textual feedback. Since \dataset{} is the first to evaluate multi-turn multi-modal code generation, we attribute this gap to the superior generalization capabilities of proprietary models, an area where contemporary open-source VLMs still fall short.
    \item \textbf{Forgetting issue:} The multi-turn setting introduces a significant forgetting issue, causing previously implemented functions or design elements to be contradicted or overwritten by later code. This issue consistently occurs across all leading models and results in a performance degradation of up to 46\%, as measured by the \textit{forgetting rate} metric reported in Table \ref{tab:main}.
    \item \textbf{Challenges of visual feedback}: Visual feedback proves consistently more difficult than textual feedback, resulting in significantly lower pass rate. Figure \ref{fig:visual_error} shows a detailed breakdown of failure modes, with most common ones being (1) the literal replication of layout sketches without implementing the underlying functionality, and (2) overlooking textual annotations in densely annotated images.
    \item \textbf{Key factors affecting usability}: The determinants of usability evolve with model capability, as discussed in \S\ref{sec:results:eval_analysis}. For weaker models, poor usability is primarily linked to \textit{broken or non-functional features}; However, for stronger models that achieve high pass rates, usability is usually impacted by \textit{design flaws}, such as inefficient navigation or a lack of interactive feedback. This highlights that achieving high usability requires implementing implicit, common-sense design principles that are often omitted from user instructions.%
\end{enumerate}

To address the critical forgetting issue, we propose \method{} as a simple yet strong baseline using \textbf{\textit{a}}gent \textbf{\textit{c}}ritique to \textbf{\textit{e}}nhance user instructions. %
For each turn in the conversation, \method{} begins with the LLM generating an \textit{initial code solution} based on user instruction. Then, a web agent, powered by the same LLM, autonomously explore the rendered website to verify the implementation of both the current and past instructions. Coding failures or oversights identified by the agent are appended to the user instruction, prompting the model to explicitly prevent these specific pitfalls in the \textit{secondary coding attempt}. Experiments show that \method{} reduces forgetting rate to \textbf{nearly zero} and significantly improves the pass rate by up to \textbf{9.3\%} with textual feedback. However, the performance gains are less significant with visual feedback, leading to gains up to \textbf{5.9\%}, further highlights the challenge of effective code generation based on visual feedback.

In summary, our contributions are as follows: (1) We propose \dataset{}, a realistic benchmark for \textit{multi-turn} front-end coding with both \textit{textual} and \textit{visual} feedback; (2) We identify critical limitations of current models, including significant performance degradation caused by \textit{forgetting issue} and a significant difficulty in \textit{interpreting visual feedback}, especially for open-source VLMs; (3) We propose \method{}, an agent-based critique method that effectively mitigates the forgetting issue especially with textual feedback.%

\section{\dataset{}}

\subsection{Task Formulation}
\label{sec:dataset:task_formulation}

\newcommand{\intent}[0]{\ensuremath{\mathbf{i}}}
\newcommand{\inst}[0]{\ensuremath{\tilde{\mathbf{i}}}}
\newcommand{\user}[0]{\textsc{SimUser}}

\textbf{Data composition.} Each data point in our dataset represents a multi-turn dialogue, consisting of a sequence of user intents $\{\intent{}_t\}_{t=1}^T$, where $T$ represents the number of turns\footnote{$T$ is set to 10 in \dataset{} dataset.}. To enable a granular evaluation of instruction following, each intent $\intent{}$ is further paired with a set of test cases $\mathcal{C}(\intent{}) = \{c_m\}_{m=1}^M$. These test cases decompose the paragraph-long intent into a series of individual, verifiable statements that form the basis for our evaluation. For example, the intent~~\textit{\color{purple}implement visual badges \uline{next to thread titles}$_\text{\hspace{1pt}(1)}$ to \uline{indicate their status with distinct colors and shapes}$_\text{\hspace{1pt}(2)}$}~~is broken down into test cases like: (1) A badge is displayed next to each thread title, and (2) Badges for different statuses are visually distinct. A detailed data example is in \S\ref{sec:appendix:data_details}.

\begin{wrapfigure}{r}{0.45\textwidth}
  \vspace{-24pt}
  \begin{minipage}{0.44\textwidth}
    \begin{algorithm}[H]
      \caption{Inference pipeline.}
      \label{alg:formulation}
      \begin{algorithmic}[1]
        \Require Model $\mathcal{M}$; user intents $\{\intent{}_t\}_{t=1}^T$; user simulator \user{}
        \Ensure website \(w\)
        \State $\mathbf{H}_1 \gets [\,]$
        \For{$t \gets 1$ \textbf{to} $T$}
\State $\inst{}_t \gets \user{}(\intent{}_t|\mathbf{H}_t)$
\State $o_t \gets \mathcal{M}(\inst{}_t | \mathbf{H}_t)$
\State $\mathbf{H}_{t+1} \gets \mathbf{H}_t \mathbin{\|} \langle \inst{}_t, o_t \rangle$
\EndFor
\State $w = \textsc{Render}(o_T)$
\State \Return $w$
      \end{algorithmic}
    \end{algorithm}
  \end{minipage}
  \vspace{-12pt}
\end{wrapfigure}
\textbf{Inference pipeline.} The pipeline for evaluating a conversational model $\mathcal{M}(\cdot)$ is detailed in Alg. \ref{alg:formulation}. At each turn $t$, model $\mathcal{M}$ produces an output code $o_t$ based on both the current instruction and the full conversation history $\mathbf{H}_t$. To capture the dynamic nature of multi-turn conversations, where later turns are conditioned on prior interactions, the static intent $\intent{}$ is not used directly as model input. Instead, we employ a user simulator, $\user{}(\cdot)$, to transform the static intent $\intent{}_t$ into a context-aware instruction $\tilde{\mathbf{i}}_t = \user{}(\mathbf{i}_t | \mathbf{H}_t)$, which will serve as the actual input to the model. The final website $w$, rendered from the last turn's output code $o_T$, is then used for  evaluation.

\textbf{Evaluation.} To comprehensively evaluate the quality of the generated websites, we assess two key dimensions: \textit{instruction following} and \textit{user experience}.

The evaluation of instruction following relies on the predefined set of test cases $\mathcal{C}(\intent{})$. We define the \textbf{pass rate} (PR) as our primary metric for implementation correctness. This is derived by first calculating the pass count (PC) -- the number of test cases passed for a given intent \intent{} -- and then aggregating these counts across all intents for the final code output $o_T$:
\begin{align}
    \text{PC}(o | \intent{}) = \sum_{c\in\mathcal{C}(\intent{})} \mathbf{I}\left(\textsc{Render}(o)\hspace{4pt}\text{passes}\hspace{4pt}c\right);\quad   
    \text{PR}(o) = \frac{\sum_\intent{} \text{PC}(o|\intent{})}{|\sum_\intent{}\mathcal{C}(\intent{}) |}
\end{align}
In practice, the pass count is measured by a web agent that interacts with the website and verifies each test case, as in \S\ref{sec:dataset:agent_eval}.
Motivated by observations that features implemented in earlier turns are often overwritten in later turns, we further introduce the \textbf{forgetting rate} (FR) metric. It measures how much functionality correctly implemented in prior turns ($t<T$) is lost in the final output:
\begin{align}
\text{FR} = 1 - \frac{\sum_{t=1}^{T-1}\text{PC}(o_T|\intent{}_t)}{\sum_{t=1}^{T-1} \text{PC}(o_t|\intent{}_t)}
\end{align}

While these metrics assess implementation correctness when following user instructions, they may not fully capture the overall user experience. To address this, we also evaluate \textbf{usability}, defined as how easily a new user can learn to navigate the website and accomplish tasks. We evaluate this by deploying an LLM agent that simulates a first-time user with no prior knowledge of the website. The agent autonomously explores the interface, formulates tasks based on its observations, and attempts to complete them, yielding a behavioral trajectory. We then employ a \textit{pairwise comparison protocol}, where an LLM judge reviews trajectories from two different websites and determines which provides a superior user experience. The final metric is the win rate against a curated set of reference websites, ensuring a robust and standardized evaluation.

\textbf{Further details.} Additional details are provided in subsequent sections: data curation in \S\ref{sec:dataset:curation}, user simulator in \S\ref{sec:dataset:user_simulation}, and the agent-based evaluation in \S\ref{sec:dataset:agent_eval}.

\subsection{Data Curation}
\label{sec:dataset:curation}

\begin{figure}[!b]
\vspace{-10pt}
    \centering 
    \begin{minipage} {0.48\textwidth}
        \centering
        {\includegraphics[width=\textwidth]{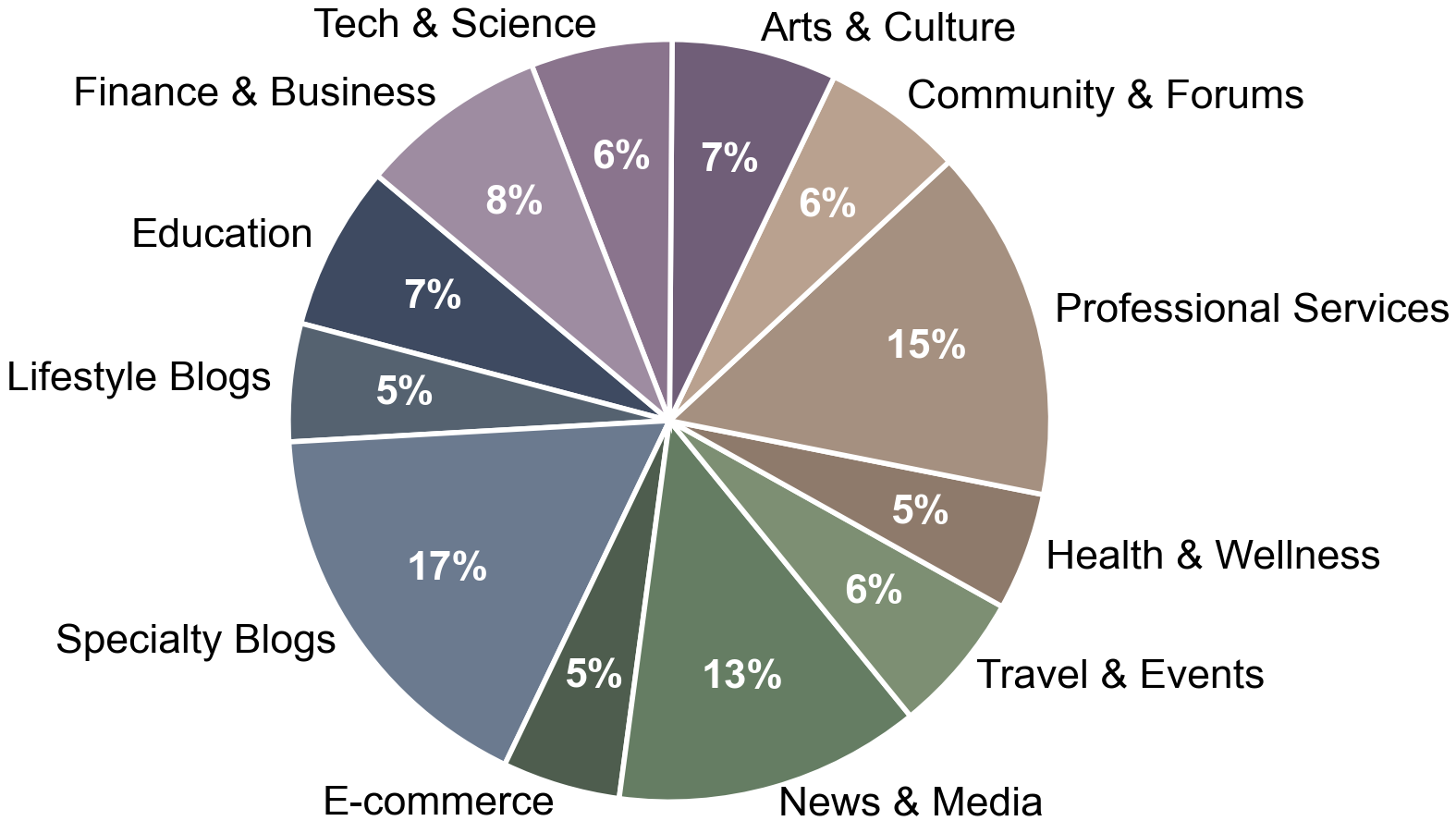}}
        \caption{\small Website summary topics distribution.}
        \label{fig:data:summary}
    \end{minipage}
         \hfill
        \begin{minipage} {0.48\textwidth}
        \centering
        \begin{tabular}{l|l}
            \hline
            \multicolumn{2}{c}{\textit{Dataset Size}} \\
            \hdashline
            \# Dialogues & 100 \\
            \# Turns & 1,000 \\
            \# Test Cases & 3,676 \\
            \hline
            \multicolumn{2}{c}{\textit{User Intent Length}} \\
            \hdashline
            \# Words per Turn & 93.7 \\
            \# Words per Dialogue & 936.8 \\
            \hline
        \end{tabular}
        \captionof{table}{\small Statistics of curated user intents and test cases in \dataset{}.}
        \label{tab:statistics}
    \end{minipage} 
\end{figure}

\textbf{Source website selection.}
To curate a diverse and challenging set of user intents reflective of real-world scenarios, we begin by sampling 10,000 webpages from the C4 dataset \citep{raffel2020exploring}. For each site, we generate summaries of its content and key features using GPT-4o. We then apply BERTopic \citep{grootendorst2022bertopic} to these summaries to identify thematic clusters. From the most prominent clusters, we manually select 100 websites --  one representative site per cluster -- while excluding low-information pages (e.g., placeholders). This process yields a realistic and diverse set of seed websites spanning a wide range of domains, as summarized in Figure \ref{fig:data:summary}.

\textbf{Automatic data generation.} Using the curated website summaries, we prompt GPT-4o to generate user intents $\intent{}_t$ and their paired test cases $\mathcal{C}(\intent{}_t)$ for each conversational turn. These intents guide the subsequent user simulation (details in \S\ref{sec:dataset:user_simulation}). We categorize user intents into two types: \textit{functionality} (e.g., navigation, form submission, search) and \textit{design} (e.g., layout, colors, images, visual effects). For each turn, we randomly select an intent type and prompte GPT-4o to generate a corresponding user intent along with a set of verifiable test cases. Figure \ref{fig:data:intent_wordcloud} shows the distribution of words in the generated intents for each type, illustrating their diversity.

\textbf{Manual refinement and validation.} To ensure the reliability of our evaluation, we manually refined all automatically generated test cases to ensure that they are hallucination-free, unambiguous, and verifiable. We removed entirely unsatisfactory test cases and edited those that were partially satisfactory. This refinement process led to the removal of 66 test cases (1.7\%) and the editing of 2,783 test cases (73.4\%). To validate the improvement, a secondary independent annotator reviewed a sample of 224 test cases both before and after editing. The manual refinement increased the validity rate of test cases from 90.1\% to \textbf{94.7\%}. Statistics of the final dataset are in Table \ref{tab:statistics}.

\subsection{User Simulation}
\label{sec:dataset:user_simulation}

\begin{figure}[!t]
    \centering
    \includegraphics[width=\linewidth]{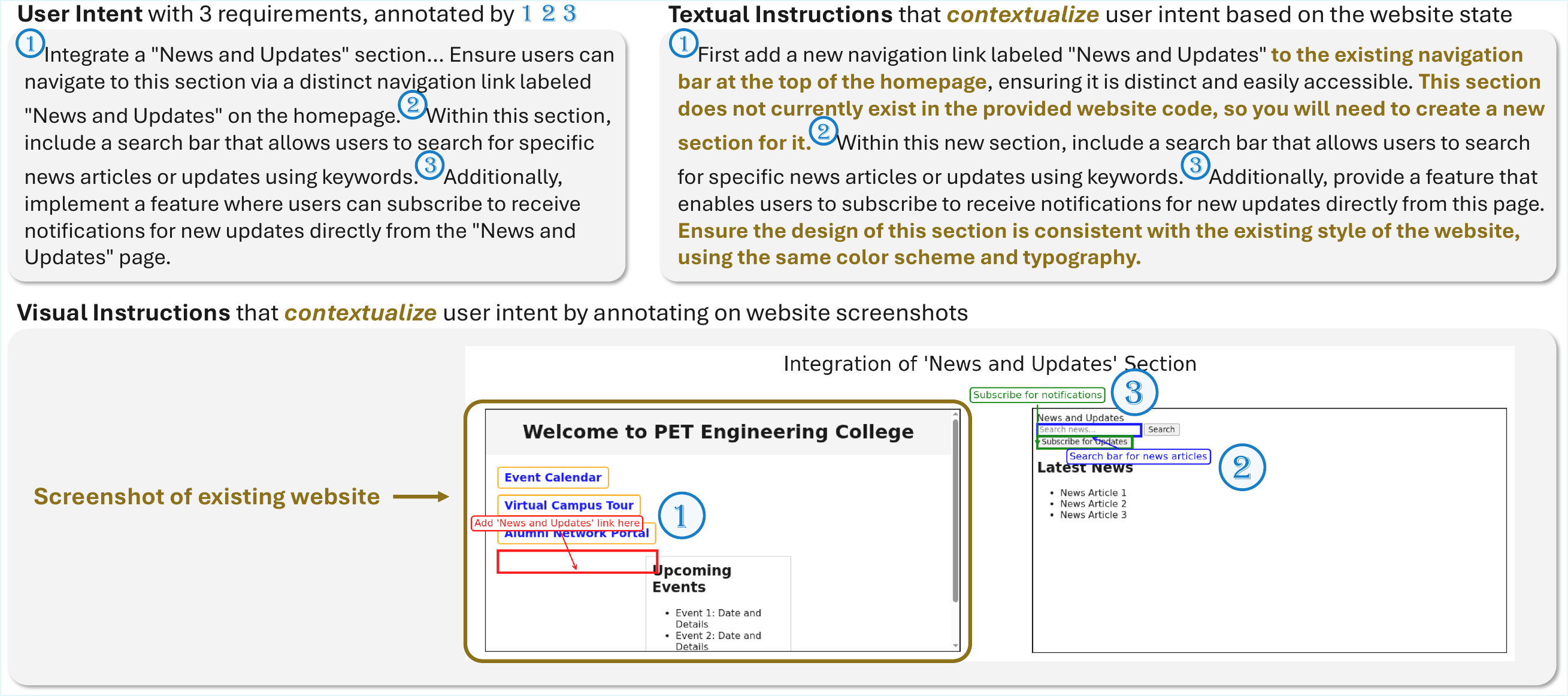}
    \caption{\small\textbf{Illustration of the user simulators}. Both simulators aim to contextualize the user intent based on the current website state: The textual simulator rewrites the textual content, while the visual simulator draws sketches and annotations on top of website screenshots.}
    \label{fig:user_sim}
    \vspace{-12pt}
\end{figure}

The {user intents} are curated based on the assumption of an ideal website state. However, the actual website generated during a multi-turn conversation may deviate from this assumption, creating a contextual gap when intents are applied directly as model inputs. The objective of user simulation is to bridge this gap by dynamically adapting static {user intents} into context-aware \textit{user instructions}, preserving the semantic core of the original intent while grounding it in the current state of the model-generated website. Specifically, user simulators perform two critical functions: (1) \textbf{Resolving ambiguity}: When an intent refers to an element that is ambiguous or absent on the current website, the simulator refines the reference (e.g., ``the submit button'' $\rightarrow$ ``the submit button on the right") or requests the creation of the missing element. (2) \textbf{Omitting redundancy}: If an intent requests a feature that has already been implemented, the simulator adapts the instruction to avoid redundant requests (e.g., ``implement a navigation bar" $\rightarrow$ ``modify the existing navigation bar").

Following these principles, we implement two types of user simulators as in Figure \ref{fig:user_sim}:
\begin{enumerate}[leftmargin=*, itemsep=0pt, topsep=0pt]
    \item \textbf{Textual instruction simulator}: An LLM is provided with the current code state and the original user intent, and is instructed to rewrite the intent into a textual user instruction.
    \item \textbf{Visual instruction simulator}: A VLM is equipped with \textit{drawing tools} and provided with screenshots of the current website. It produces sketches, shapes, and text annotations to anchor instructions directly onto the visual interface. This simulator is permitted multiple turns of drawing to fully articulate the user intent.
\end{enumerate}
Implementation details and simulated instruction examples are presented in \S\ref{sec:appendix:user_simulator}. We validate the reliability of our proposed user simulation with two metrics: (1) \textit{intent preservation}, measured by the ratio of test cases well represented by the user instructions, and (2) \textit{contextual faithfulness}, as measured by the ratio of user instructions that align well with the actual website state. A manual review of 114 simulated instructions and 411 test cases shows that textual simulators achieve high intent preservation (98\%) and strong contextual faithfulness (96\%), proving to be highly reliable. Visual simulators also exhibit high contextual faithfulness (97\%) due to grounding user intents directly in rendered screenshots, but demonstrate much lower intent preservation (76\%) owing to limited tool-use effectiveness in GPT-4o. Despite this limitation, we view visual simulation as a valuable proof of concept: with our benchmark and more capable future models, the reliability of visual user simulation can be further improved.

\subsection{Agent-Based Evaluation}
\label{sec:dataset:agent_eval}

Existing evaluation approaches in front-end development largely rely on static artifacts like rendered screenshots or source code \citep{si2025design2code,sun2025fullfront}. However, both approaches have fundamental limitations. For example, consider a dynamic test case: \textit{\color{purple}selecting an option from the dropdown menu dynamically updates the page}. Screenshot-based metrics are not capable of verifying such cases, since they capture only a single visual state and cannot track dynamic user interactions. Meanwhile, code-based analysis, while theoretically more comprehensive, often misaligns with the end-user experience and may overlook subtle logical errors. In this example, the dropdown's filtering logic might appear syntactically correct and structurally sound, leading a code-based evaluation to register a false positive. Yet, a subtle bug could prevent it from functioning as intended -- an error that only becomes apparent during actual user interaction.

To overcome these shortcomings and better emulate authentic user assessment, we employ an \textbf{interactive web agent} that evaluates the website through direct interaction. Our agent-based evaluator operates in two modes for the two different metrics: for \textit{pass rate}, it acts as an informed expert, using the full context to verify a test case, while for \textit{usability}, it simulates a naive user, exploring the site and attempting self-proposed tasks. We adapt the WebVoyager \citep{he2024webvoyager} framework where an agent receives annotated screenshots as input, and performs actions such as clicking and scrolling chosen from a predefined action space. We additionally equip the agent with a novel suite of \textit{image manipulation tools} such as cropping UI elements and comparing visual states to enhance the agent's visual perception. These tools improve alignment with human judgments on design-centric tasks involving fine-grained visual details, such as verifying a subtle hover effect in a local region, as in Table \ref{tab:pr_iaa} and Figure \ref{fig:imtool}. Implementation details are in \S\ref{sec:appendix:agent}.

\section{Experiments}

We evaluate 20 models from nine model families, spanning proprietary models like \raisebox{-0.15em}{\includegraphics[height=1em]{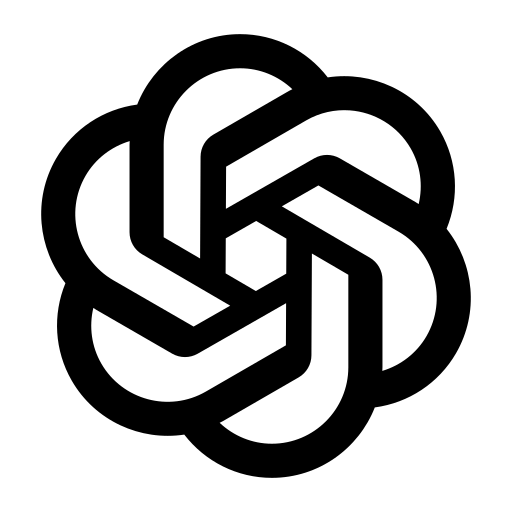}}\hspace{1pt}OpenAI's GPT, \raisebox{-0.1em}{\includegraphics[height=1em]{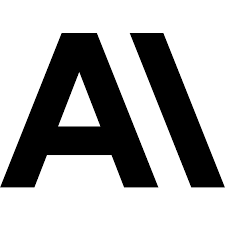}}\hspace{1pt}Anthropic's Claude, \raisebox{-0.1em}{\includegraphics[height=1em]{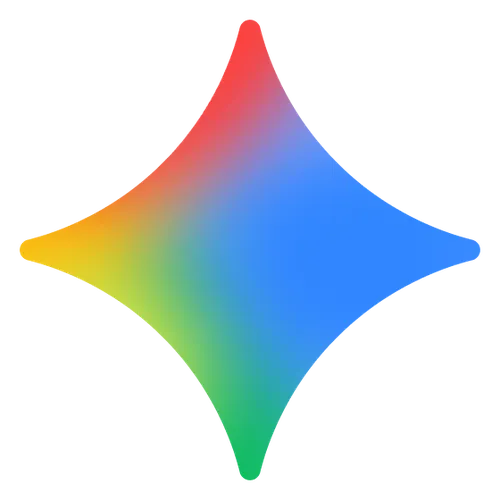}}\hspace{1pt}Google's Gemini, as well as open-source models like \raisebox{-0.1em}{\includegraphics[height=1em]{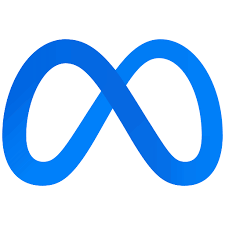}}\hspace{1pt}Meta's Llama, \raisebox{-0.1em}{\includegraphics[height=1em]{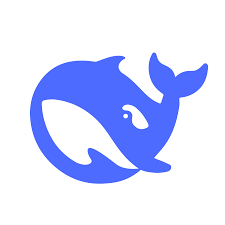}}\hspace{1pt}DeepSeek-R1 \citep{guo2025deepseek}, \raisebox{-0.1em}{\includegraphics[height=1em]{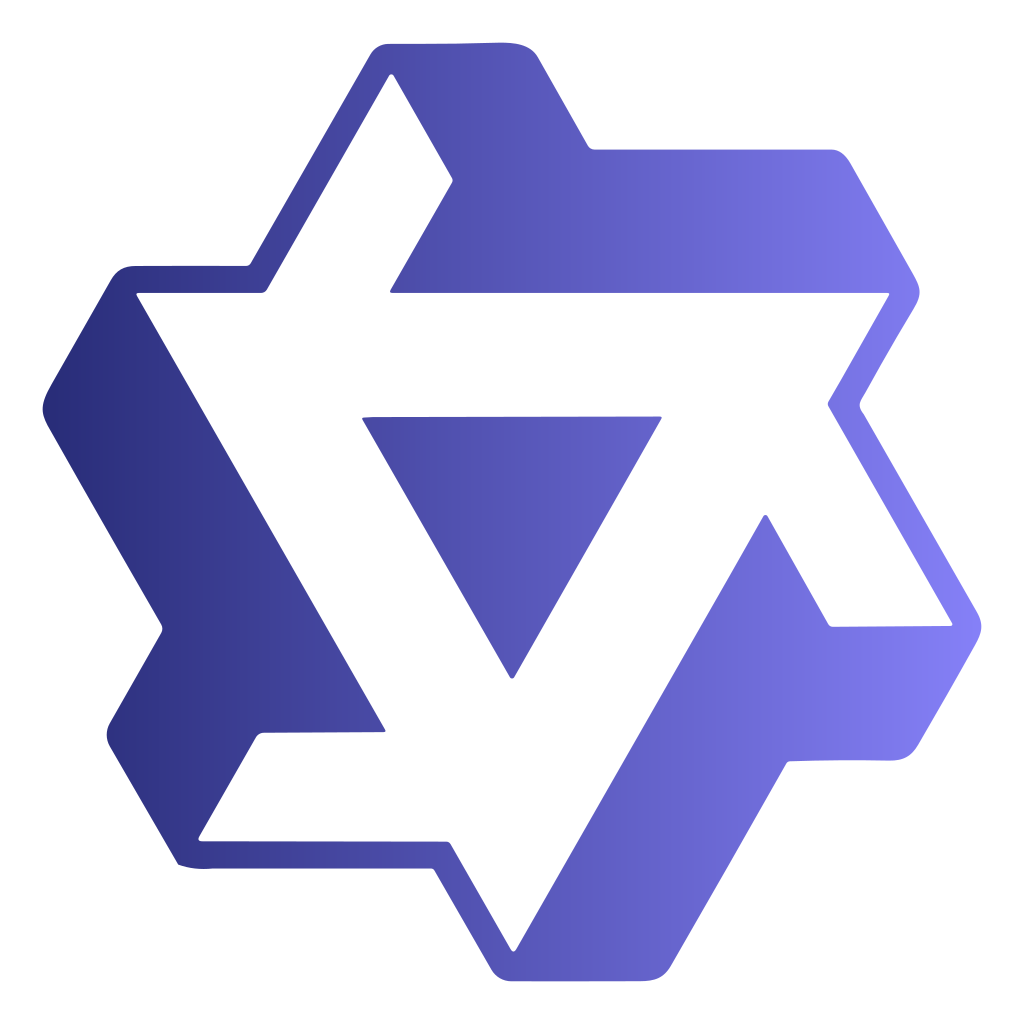}}\hspace{1pt}Qwen \citep{qwen2.5-VL,qwen3technicalreport},
\raisebox{-0.15em}{\includegraphics[height=1em]{figures/gpt_icon.png}}\hspace{1pt}OpenAI's GPT-OSS,
\raisebox{0em}{\includegraphics[height=.7em]{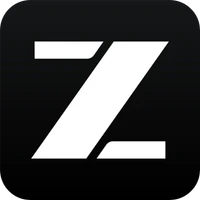}}\hspace{1pt}GLM \citep{hong2025glm} and \raisebox{-0.1em}{\includegraphics[height=.9em]{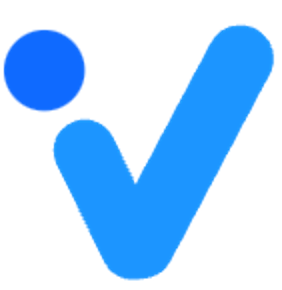}}\hspace{1pt}Ovis \citep{lu2025ovis25technicalreport}.
For evaluation, we employ the web agent powered by GPT-4o to perform agent-based evaluation. We report \textit{pass rate} (PR) and \textit{usability} (UX) as main metrics and \textit{forgetting rate} (FR) as a supplementary metric to measure the forgetting issue.

\subsection{Main Results}
\label{sec:results}

\begin{table}[!t]
    \centering
    \begin{tabular}{l@{\hspace{.8em}}l|rr|rrr|rrr}
    \hline
      \rowcolor{gray!10}  &  & \multicolumn{2}{c|}{{Single-Turn (T)}} & \multicolumn{3}{c|}{{Multi-Turn (T)}} & \multicolumn{3}{c}{{Multi-Turn (V)}} \\
      \rowcolor{gray!10}   & & {PR$\uparrow$} & {UX$\uparrow$} & {PR$\uparrow$} & {FR$\downarrow$} & {UX$\uparrow$} & {PR$\uparrow$} & {FR$\downarrow$} & {UX$\uparrow$} \\
    \hline
    \rowcolor{gray!5}  \multicolumn{10}{c}{\textit{Proprietary VLMs}} \\
    GPT-4o &- &  51.4 & 50.0 & {56.0} & 21.4 & {55.0} & 55.0{\scriptsize\color{red}$\downarrow$1.0} & 8.0 & 52.5{\scriptsize\color{red}$\downarrow$2.5} \\
    Claude-4-Sonnet &-& \textbf{77.5} & \textbf{71.8} & 45.5 & 22.8 & 60.8 & 59.3{\scriptsize\color{mygreen}$\uparrow$13.8} & 10.6 & 64.8{\scriptsize\color{mygreen}$\uparrow$4.0} \\
    Gemini-2.5-Pro &- & {57.2} & {56.8} & \textbf{75.0} & \textbf{4.5} & \textbf{71.0} & \textbf{68.7}{\scriptsize\color{red}$\downarrow$6.3} & \textbf{6.2} & \textbf{73.8}{\scriptsize\color{mygreen}$\uparrow$2.8} \\
    \hline
   \rowcolor{gray!5}   \multicolumn{10}{c}{\textit{Open-Source LLMs}} \\
    Qwen3 &8B & 54.3 & 45.8 & 59.9 & 13.9 & 57.8 & n/a & n/a & n/a \\
    GPT-OSS&20B & 64.3 & {68.5} & {61.6} & 15.8 & \textbf{71.5}& n/a & n/a & n/a \\
    Qwen3-Coder&30B & 54.1 & 49.8 & 61.0 & \textbf{8.6} & 54.3& n/a & n/a & n/a \\
    Llama-3.3 & 70B & 38.4 & 35.8 & 28.9 & 46.3 & 40.8& n/a & n/a & n/a \\
    GPT-OSS & 120B & \textbf{68.0} & \textbf{73.5} & 57.9 & {9.2} & {66.8}& n/a & n/a & n/a \\
    Qwen3&235B & 62.9 & 57.8 & 59.0 & 20.6 & 61.3& n/a & n/a & n/a \\
    Qwen3-Coder&480B & {67.0} & 64.0 & \textbf{62.5} & 15.4 & 62.3& n/a & n/a & n/a \\
    DeepSeek-R1 & 685B & 60.0 & 53.5 & 24.7 & 34.5 & 40.0 & n/a & n/a & n/a \\
    \hline
   \rowcolor{gray!5}   \multicolumn{10}{c}{\textit{Open-Source VLMs}} \\
    Qwen2.5-VL & 7B & 24.5 & 19.5 & 23.7 & 28.2 & 26.8 & 16.1{\scriptsize\color{red}$\downarrow$7.6} & 39.3 & 17.0{\scriptsize\color{red}$\downarrow$9.8} \\
    GLM-4.1V-Thinking & 9B & 36.6 & 32.3 & 18.2 & 5.8 & 41.5 & 15.3{\scriptsize\color{red}$\downarrow$2.9} & \textbf{4.3} & 32.0{\scriptsize\color{red}$\downarrow$9.5} \\
    Ovis2.5 & 9B  & 51.5 & 49.3 & {35.3} & 12.2 & 34.3 & 27.0{\scriptsize\color{red}$\downarrow$8.3} & 10.5 & 39.5{\scriptsize\color{mygreen}$\uparrow$5.2} \\
    Gemma3 & 12B & 47.1 & 43.5 & 43.8 & 30.3 & 43.0 & 30.4{\scriptsize\color{red}$\downarrow$13.4} & 22.1 & 38.5{\scriptsize\color{red}$\downarrow$4.5} \\
    Gemma3 & 27B & 36.7 & 37.2 & 37.2 & 42.1 & 39.5 & 35.5{\scriptsize\color{red}$\downarrow$1.2} & 16.7 & \textbf{41.0}{\scriptsize\color{mygreen}$\uparrow$2.5} \\
    Qwen3-VL & 30B & \textbf{54.6} & \textbf{50.0} & \textbf{54.1} & \textbf{4.7} & \textbf{50.3} & 39.1{\scriptsize\color{red}$\downarrow$15.0} & 12.3 & 35.8{\scriptsize\color{red}$\downarrow$14.5} \\
    Qwen2.5-VL & 32B & 44.5 & 38.0 & 33.2 & 46.9 & 37.5 & 36.9{\scriptsize\color{mygreen}$\uparrow$3.7} & 25.7 & 36.5{\scriptsize\color{red}$\downarrow$1.0} \\
    Qwen2.5-VL & 72B & 43.0 & 33.8 & 52.8 & 17.8 & 46.5 & \textbf{44.6}{\scriptsize\color{red}$\downarrow$8.2} & 15.4 & 37.3{\scriptsize\color{red}$\downarrow$9.2} \\
    GLM-4.5V & 108B & 24.4 & 23.5 & 35.2 & 9.8 & 40.5 & 7.9{\scriptsize\color{red}$\downarrow$27.3} & 27.7 & 7.8{\scriptsize\color{red}$\downarrow$32.7} \\
    \hline
    \end{tabular}
    \vspace{-4pt}
    \caption{\small \textbf{Main evaluation results.} We report the pass rate (PR$\uparrow$) and usability (UX$\uparrow$) as main metrics, and forgetting rate (FR$\downarrow$) to additionally quantify the forgetting issue. The best performance in each block is \textbf{bolded}. For model performance with visual feedback (V), we report their difference against the performance with textual feedback (T) represented as {\scriptsize\color{red}$\downarrow$} or {\scriptsize\color{mygreen}$\uparrow$}.}
    \label{tab:main}
    \vspace{-12pt}
\end{table}

The main results are shown in Table \ref{tab:main}. Overall, \textbf{all models perform substantially below perfect accuracy} on our challenging \dataset{} dataset. Even the state-of-the-art model, Gemini-2.5-Pro, falls short of perfect performance by 25.0\% with textual feedback and 31.3\% with visual feedback. The gap is more significant for open-source models, where the top models achieve only 62.5\% with textual instructions and 44.6\% with visual instructions. Notably, \textbf{the gap between closed-source and open-source model is more pronounced with visual feedback}. Since many models are heavily tuned for code generation from text, but not for code generation from image, we attribute the superior performance of proprietary models to stronger generalization capabilities. Our findings thus identify poor generalization as a key shortcoming of current open-source VLMs and an important direction for future work.

\textbf{Forgetting issue.} Table \ref{tab:main} reveals a significant forgetting issue across all models, with their forgetting rate ranging from 4.3\% to 44.6\%.
The forgetting issue typically occurs when instructions from multiple turns request modifications to the same component of code (e.g., adding different buttons to the same section). Although these instructions are not mutually exclusive, the model is at risk of overwriting the previous implementation instead of integrating the new feature, as exemplified in Figure \ref{fig:forgetting_example}. Mitigating this issue requires models to accurately recall past instructions and dynamically navigate the codebase to preserve existing functionalities while implementing new ones -- a capability that proves challenging for current models.

\textbf{Visual v.s. textual instructions.} As we evaluate VLMs with either textual or visual instructions, we are able compare their performance across the two settings. While the two instruction types are designed to convey the same set of user intents, Table \ref{tab:main} shows a consistent performance degradation for most VLMs when processing visual instructions, especially for open-source VLMs. To identify the  challenges specific to visual
\begin{wrapfigure}{r}{0.6\textwidth} 
        \centering
    \includegraphics[width=\linewidth]{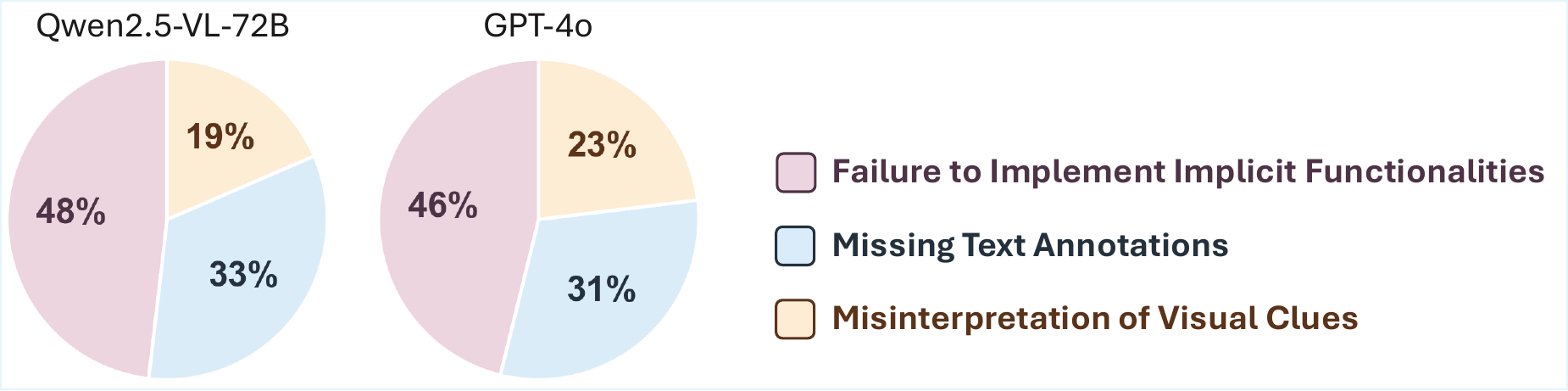}
    \vspace{-6pt}
    \caption{\small Error breakdown in interpreting visual instructions.}
    \label{fig:visual_error} 
    \vspace{-12pt}
\end{wrapfigure}
instruction following, we manually analyze 608 test cases from the outputs of Qwen2.5-VL-72B and GPT-4o. Our analysis focus on the \textit{subset} of instances where a model fails a visual instruction but succeeds on its textual equivalent, thereby isolating errors caused by visual misinterpretation. We categorize these errors into three primary types, as shown in Figure \ref{fig:visual_error}: (1) \textbf{Failure to implement implicit functionalities}, the most common error, where VLMs replicate the UI layout but fail to implement the underlying functionalities (e.g., failing to implement the page-update behavior for navigation arrows, as in Figure \ref{fig:error1}); (2) \textbf{Missing text annotations}, where model miss some text annotations in the image, especially in densely annotated images (Figure \ref{fig:error2}); and (3) \textbf{Misinterpretation of visual clues}, the least common error, which occurs when visual clues are slightly ambiguous (Figure \ref{fig:error3}).

\textbf{Multi-turn v.s. single-turn.} To isolate the challenges specific to multi-turn interactions and assess the inherent task complexity, we perform a single-turn baseline by concatenating all user intents. As shown in Table \ref{tab:main}, most models still fail to exceed a 70\% pass rate in this setting, confirming the inherent difficulty of our task. Compared to the single-turn setting, the impact of multi-turn interactions is model-dependent: while some models (e.g., GPT-4o) benefit from the step-by-step task decomposition, others (e.g., Claude-4-Sonnet) are penalized due to forgetting and long-context challenge. As shown in Figure \ref{fig:gets_worse_issue}, certain models (e.g., Claude-4-Sonnet and DeepSeek-R1) are particularly susceptible to long-context challenges: Although these models initially achieve high pass rates comparable to GPT-4o and Gemini-2.5-Pro, their performance degrades significantly in later turns as the context grows. This highlights effective long-context handling as another key challenge.

\subsection{Analysis for Automatic Evaluation}
\label{sec:results:eval_analysis}

To validate the reliability of our automatic evaluation metrics, we conduct human evaluation studies to measure the alignment between the automatic evaluation results and human judgments.

\begin{figure}[!t]
    \centering 
    \begin{minipage}  {0.43\textwidth}
    \includegraphics[width=\linewidth]{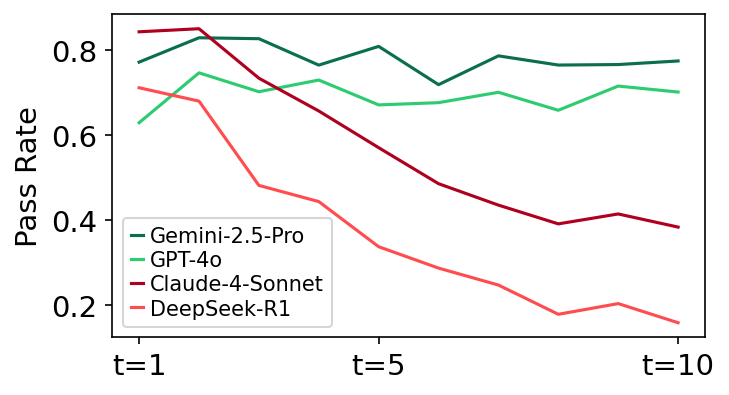}
    \vspace{-16pt}
    \caption{\small \textbf{Pass rate across different turns.} We show the pass rate of intermediate outputs, i.e. PR$(o_t|\mathbf{i}_t)$, to factor out the forgetting issue.}%
    \label{fig:gets_worse_issue} 
    \vspace{-10pt}
    \end{minipage}
    \hfill
\begin{minipage}   {0.55\textwidth}
\small
\centering
\setlength{\tabcolsep}{4pt}
\begin{tabular}{l|rr|rr|rr}
\hline
\rowcolor{gray!10} & \multicolumn{2}{c|}{\scriptsize All Types} & \multicolumn{2}{c|}{\scriptsize Design} & \multicolumn{2}{c}{\scriptsize Functionality} \\[-0.1em]
\rowcolor{gray!10} & \multicolumn{1}{c}{$acc$} & \multicolumn{1}{c|}{$\kappa$} & \multicolumn{1}{c}{$acc$} & \multicolumn{1}{c|}{$\kappa$} & \multicolumn{1}{c}{$acc$} & \multicolumn{1}{c}{$\kappa$} \\
\hline
\color{gray} Random evaluator & \color{gray}50.4 & \color{gray}0.9 & \color{gray}52.1 & \color{gray}4.2 & \color{gray}48.4 & \color{gray}0.0 \\
\color{black} Image evaluator & 70.4 & 39.1 & 70.0 & 37.9 & 72.1 & 37.9 \\
Code evaluator & 72.5 & 42.1 & 68.9 & 37.5 & 76.7 & 46.0 \\
\hline
\textbf{Agent evaluator} & \textbf{82.0} & \textbf{62.7} & \textbf{81.2} & \textbf{62.3} & \textbf{82.9} & \textbf{62.4} \\
\quad w/o IM tools &81.4 & 61.8 & 80.3 & 60.6 & 82.7 & 62.1 \\[-0.35em]
   & {\scriptsize\red{-0.6}} & {\scriptsize\red{-0.9}} & {\scriptsize\red{-0.9}} &  {\scriptsize\red{-1.7}} & {\scriptsize\red{-0.2}} & {\scriptsize\red{-0.3}} \\
\hline
\end{tabular}
\vspace{-5pt}
\captionof{table}{\small \textbf{Human alignment for the automatic evaluation of pass rate}, measured by agreement accuracy ($acc$) and Cohen's Kappa ($\kappa$). We compare our evaluator against image-based evaluator, code-based evaluator, and an ablated version without image manipulation tools (w/o IM tools).}
\label{tab:pr_iaa}
        \vspace{-10pt}
\end{minipage}
\end{figure}

\textbf{Pass rate.} We manually evaluate 218 test cases against two sets of model outputs. As shown in Table \ref{tab:pr_iaa}, our agent-based evaluator demonstrates significant alignment with human annotators, achieving an accuracy of 82.0 and a Cohen's Kappa of 62.7. In contrast, the image-based evaluator cannot handle dynamic test cases. The code-based evaluator, while theoretically capable of judging all cases, tends to be overly optimistic and misses subtle errors, leading to weaker alignment and a higher false positive rate (19.0\%, compared to 12.2\% for our agent-based evaluator). We conduct an ablation study by removing our proposed image manipulation tools from the agent-based evaluator (w/o IM) and measure their resulted alignment scores. Results show that our full evaluator consistently achieves higher alignment, with the benefit being more pronounced for design-category test cases. Figure \ref{fig:imtool} illustrates an example where these tools are critical for evaluating fine-grained visual details.

\textbf{Usability.} We evaluate usability based on interaction trajectories generated by a web agent simulating a first-time user. These trajectories exhibit plausible, human-like interaction behaviors, as illustrated by the example in Table \ref{tab:my_long_table}. To validate the reliability of automatic evaluation, we manually compare 50 trajectory pairs and assess alignment with the automated judge. We find that the automated judge aligns significantly with human evaluators, achieving an accuracy of 84.0 and a Cohen's Kappa of 66.7. Further analysis reveals that the determinants of usability shift with model proficiency: \textit{For lower-performing models, usability is mostly affected by basic functionality}, where non-functional features like broken navigation links can severely degrade the user experience. \textit{For higher-performing models, usability hinges on nuanced design choices} that are not always explicitly specified in user instructions, such as intuitive navigation (e.g., an easily accessible navigation bar) and clear system feedback (e.g., a confirmation message after a form submission). These findings highlight front-end development as a multi-dimensional challenge. Improving the user experience requires more than just correct code; it demands thoughtful, user-centric design, which is a critical direction for future research.

\subsection{Interaction with Diverse Users}

\begin{wraptable}{r}{0.48\textwidth}
    \centering
\vspace{-10pt}
    \begin{tabular}{ll|rr}
    \hline
      \rowcolor{gray!10}  Turn & User & {PR$\uparrow$} & {UX$\uparrow$} \\
    \hline
    Single & - & 51.4 & 50.0 \\
    Multi & Detailed (Standard) & \textbf{56.0} & \textbf{55.0} \\
    \hline
    Multi & Clarification-Only & 23.9 & 31.0 \\
    Multi & Preference-Only & 16.4 & 21.0 \\
    \hline
\end{tabular}
\caption{\small Performance of GPT-4o with users of different interaction types with textual instructions.}
\label{tab:interaction}
\vspace{-10pt}
\end{wraptable}
Prior sections assume that users express their intents through detailed, explicit instructions. In practice, however, users often cannot provide fully formed instructions -- either because they have not yet fully clarified their own goals, or because they find it difficult to articulate them without guidance. Consequently, LLMs must take actively elicit or infer the user intents before executing tasks. To evaluate model performance in such under-specified interactions, we simulate two user types: (1) \textbf{clarification-only user} that responds to direct questions but provides no unsolicited details, and (2) \textbf{preference-only user} that verbalizes only minimal requirements but instead reveals intent through choices among model-generated options.
These simulated users have full access to the true intent but are constrained to communicate in the specified style. All settings share the same interaction budget ($N$ turns) to ensure fairness. As shown in Table \ref{tab:interaction}, both user types significantly degrade performance compared to standard multi-turn or single-turn settings. This highlights models limited capability to actively elicit and infer user intent, directly affecting the task completion pass rate and the usability of generated websites.

    \section{\method{}: An Improved Baseline to Address Forgetting}

\begin{figure}[t]
    \centering
    \includegraphics[width=.85\linewidth]{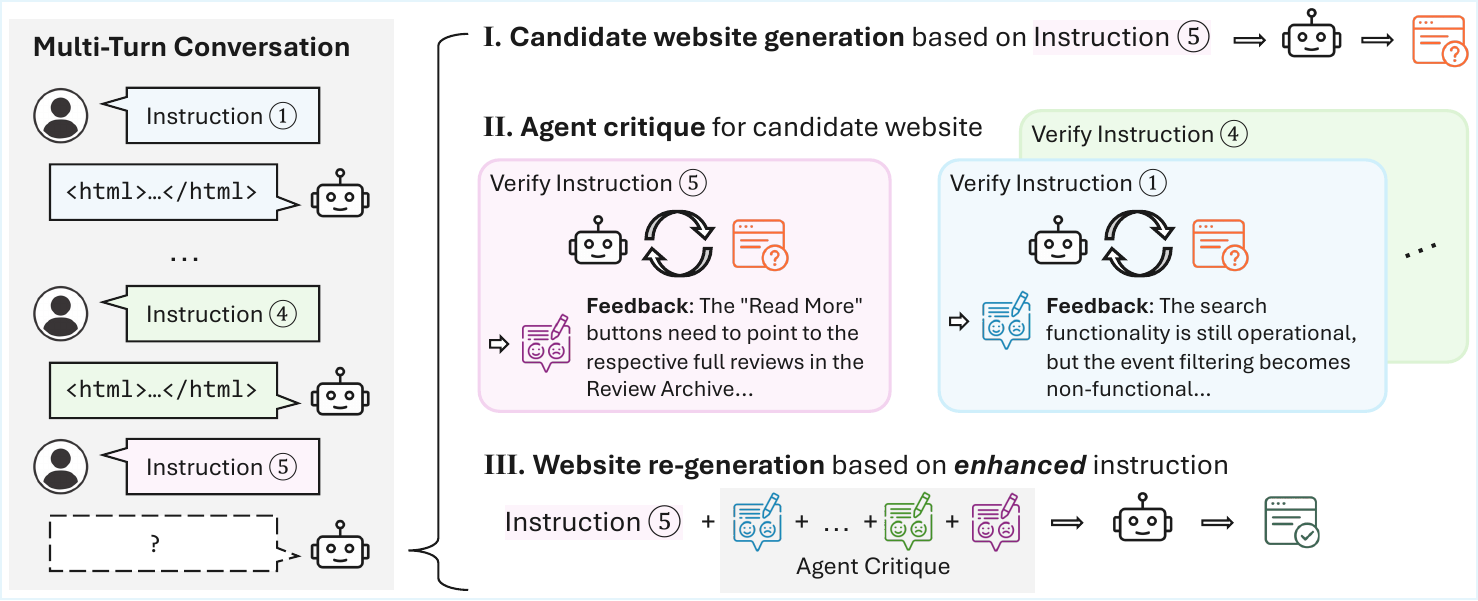}
    \caption{\small \textbf{Overview of \method{}}. \method{} first generates a candidate website, collects agent-based critique by interacting with the website and verifying the implementation of \textbf{\textit{all prior instructions}}, and then use the critique to enhance user instructions.}
    \label{fig:ours_method}
\end{figure}

\begin{table}[t]
\small
\centering
\begin{minipage}{0.54\textwidth}
    \setlength{\tabcolsep}{4pt}
    \begin{tabular}{l|rrr|rrr}
    \hline
 \rowcolor{gray!7}    &  \multicolumn{3}{c|}{Multi-Turn (T)} & \multicolumn{3}{c}{Multi-Turn (V)} \\
 \rowcolor{gray!7}   & PR$\uparrow$ & FR$\downarrow$ & UX$\uparrow$ & PR$\uparrow$ & FR$\downarrow$ & UX$\uparrow$ \\
    \hline
    GPT-4o &  56.0 & 21.4 & 55.0 & 55.0 & 8.0 & 52.5 \\
    +\method{} &  \textbf{65.3} & \textbf{0.4} & \textbf{76.3} & \textbf{60.2} & \textbf{7.2} & \textbf{65.8} \\[-0.35em]
           &  {\scriptsize\color{mygreen}+9.3} & {\scriptsize\color{mygreen}-21.0} & {\scriptsize\color{mygreen}+21.3} & {\scriptsize\color{mygreen}+5.2} & {\scriptsize\color{mygreen}-0.8} & {\scriptsize\color{mygreen}+13.3} \\[0.1em]
\hline
    Qwen2.5-VL-7B &   23.7 & 28.2 & 26.8 & 16.1 & 39.3 & 17.0 \\
     +\method{} & \textbf{25.4} & \textbf{-1.2} & \textbf{28.3} & \textbf{19.3} & \textbf{29.0} & \textbf{23.0} \\[-0.35em]
   & {\scriptsize\color{mygreen}+1.7} & {\scriptsize\color{mygreen}-29.4} & {\scriptsize\color{mygreen}+1.5} &  {\scriptsize\color{mygreen}+3.2} & {\scriptsize\color{mygreen}-10.3} & {\scriptsize\color{mygreen}+6.0}\\[0.1em]
\hline
    Qwen2.5-VL-72B &   52.8 & 17.8 & 46.5 & 44.6 & \textbf{15.4} & 37.3 \\
     +\method{} &  \textbf{58.7} & \textbf{1.5} & \textbf{51.8} & \textbf{45.4} & 17.0 & \textbf{49.0} \\[-0.35em]
    & {\scriptsize\color{mygreen}+5.9} & {\scriptsize\color{mygreen}-16.3} & {\scriptsize\color{mygreen}+5.3} & {\scriptsize\color{mygreen}+0.8} & {\scriptsize\color{red}-1.6} & {\scriptsize\color{mygreen}+11.7} \\
    \hline
    \end{tabular}
    \caption{\small \textbf{Main results of \method{}} with both textual (T) and visual (V) instructions.}
    \label{tab:ours}
\end{minipage}
\hfill
\begin{minipage}{0.42\textwidth}
    \setlength{\tabcolsep}{4pt}
\begin{tabular}{l|rrr}
\hline
 \rowcolor{gray!10} & PR$\uparrow$ & FR$\downarrow$ & UX$\uparrow$ \\
\hline
 \rowcolor{gray!5} \multicolumn{4}{c}{\textit{Baselines}} \\
Vanilla Multi-Turn & 56.0 & 21.4 & 55.0 \\
\textsc{RePrompt}& 62.3 & 6.3 & 63.8 \\
\hline
\rowcolor{gray!5} \multicolumn{4}{c}{\textit{Ablations}} \\
\textbf{\method{}} & \textbf{65.3} & {0.4} & \textbf{76.3} \\
~~- Critique Past Turns & 61.2 & 16.1 & 70.0 \\
~~- Agent Critique & 58.8 & \textbf{-0.8} & 66.3 \\
~~+ Reflection & \underline{63.4} & \underline{-0.2} & \underline{76.0} \\
\hline
\end{tabular}
\caption{\small \textbf{Ablation study of \method{}}, only evaluated on GPT-4o with textual instructions due to resource constraints.}
\label{tab:ours_ablation}
\end{minipage}
    \vspace{-8pt}
\end{table}

As discussed in \S\ref{sec:results}, even state-of-the-art models suffer from the \textbf{forgetting issue}, where features implemented in prior turns are overwritten in later ones. To address this, \method{} employs an \textbf{agent-based critique} process, detailed in Figure \ref{fig:ours_method} and Algorithm \ref{alg:ours}. First, the model generates a candidate website. A web agent then interacts with this site to verify that all features -- from both current and past instructions -- are correctly implemented. The critiques gathered during this interaction are used to augment the original instructions. Finally, this enhanced prompt is fed back to the model to \textit{regenerate} a final, improved website.

As in Table \ref{tab:ours}, \method{} consistently improves performance across all evaluated models for both textual and visual instructions. The gains are particularly significant for textual instructions, reducing the forgetting rate to nearly zero. However, improvements for visual instructions are more modest, as the agent's ability to verify feature implementation is also bottlenecked by its capacity to interpret visual instructions.

We further conduct ablation studies to validate our design choices, as in Table \ref{tab:ours_ablation}. \method{} outperforms a naive \textsc{RePrompt} baseline that re-prompts all prior instructions at each turn, as \textsc{RePrompt} may fail to detect subtle implementation errors or complex code conflicts. We further perform three additional ablations: \textbf{(1) -Critique Past Turns}: A variant that only critiques the features from the current turn improves the pass rate but fails to meaningfully reduce the forgetting rate. This confirms that proactively verifying past features is critical to mitigating forgetting. \textbf{(2) -Agent Critique}: A variant that replaces the agent critique with critique produced by a static LLM with access to the full conversation history, but not the built website. This results in a substantial performance degradation, highlighting the importance of interaction-based agent critique. \textbf{(3) +Reflection}: A variant that uses agent critiques to perform \textit{reflection} rather than \textit{regenerate} the entire website. While this variant also performs well, it is slightly inferior to our full regeneration-based approach, suggesting that complete regeneration offers greater flexibility for incorporating complex feedback.

\section{Related Work}
\label{sec:related_work}

\textbf{Front-end development.} Early research in front-end development focused on generating static webpages from images or textual descriptions \citep{gui2024vision2ui,laurenccon2024unlocking,si2025design2code,xiao2025designbench,lin2025webuibench}. Recent studies have begun to address interactivity by generating individual interactive components \citep{xiao2024interaction2code} or entire navigable sites \citep{zhu2025frontendbench,lu2025webgen}. However, they have predominantly considered a single-turn setting. Our work extends this line of research by introducing a multi-turn, conversational setting where models must build fully functional, interactive websites through iterative textual and visual feedback, moving significantly beyond existing single-turn benchmarks.

The closest prior work is is \citet{li-etal-2025-sketch2code}, which also explores multi-turn code generation from sketches. However, our benchmark differs in two critical aspects: First, the visual instructions in their work are limited to sketches, lacking the fidelity and diversity as in our work; Second,  their multi-turn setup is constrained to simplified patterns, such as progressively revealing intents or answering clarification questions. In contrast, our benchmark is designed to model the more complex and realistic conversational dynamics in iterative development.

\textbf{Multi-turn code generation.} Recent work in code generation has started to move beyond traditional single-turn benchmarks \citep{austin2021program,chen2021evaluating} toward multi-turn settings. These work simulates diverse conversational dynamics, such as users providing iterative feedback to help model complete coding tasks across multiple turns \citep{wangmint,hanconvcodeworld}, or users progressively clarifying ambiguous instructions \citep{laban2025llms}. A common limitation, however, is their exclusive focus on textual interaction, overlooking the unique dynamics of visual feedback in multi-turn interactions. To mitigate this gap, our work incorporates visual feedback, which is critical for tasks like front-end development.

\section{Conclusion}

This paper introduces \dataset{}, a novel benchmark for front-end development that models a multi-turn code generation process with multi-modal user feedback. Our benchmark employes LLM-based user simulators to dynamically adapt pre-defined intents into either textual or visual instructions. The resulted websites are evaluated based on both instruction-following correctness and overall user experience. Evaluation of 14 models reveals that current models are far from perfect and struggle with two primary challenges: (1) a significant forgetting issue, where previously implemented features are overwritten in multi-turn interactions, and (2) a persistent difficulty in interpreting visual instructions. To address the former, we propose \method{}, a strong baseline method that mitigates forgetting by leveraging a web agent to critique the implementation against past and present instructions. This work provides a solid foundation and a challenging testbed for future research in building more capable and reliable code generation approaches.

\clearpage
\newpage
\bibliographystyle{assets/plainnat}
\bibliography{paper}

\clearpage
\newpage
\beginappendix

\newcommand{\listappendixname}{}%
\newlistof{appendixtoc}{atoc}{\listappendixname}

\renewcommand{\cftappendixtocpresnum}{}
\renewcommand{\cftappendixtocaftersnum}{\hspace{0.5em}}

\newcommand{\appchapter}[1]{%
  \chapter{#1}%
  \addcontentsline{atoc}{chapter}{\protect\numberline{\thechapter}#1}%
}
\newcommand{\appsection}[1]{%
  \section{#1}%
  \addcontentsline{atoc}{section}{\protect\numberline{\thesection}#1}%
}
\newcommand{\appsubsection}[1]{%
  \subsection{#1}%
  \addcontentsline{atoc}{subsection}{\protect\numberline{\thesubsection}#1}%
}

\listofappendixtoc
\clearpage

\lstset{escapeinside={(*@}{@*)}}

\appsection{Ethics Statement}

In preparing this manuscript, we used large language models (LLMs) solely for the purpose of polishing the language (e.g., improving grammar and clarity). The LLMs were \textbf{not} employed to generate ideas, conduct analyses, provide interpretations, or alter the substance of the work. All conceptualization, methodology, data analysis, and interpretation were performed entirely by the authors. The use of the LLM was restricted to stylistic refinement, ensuring that the meaning and originality of the content remain unchanged.

\appsection{Dataset Details}
\label{sec:appendix:data_details}

\begin{figure}[!h]
    \centering
    {\includegraphics[width=\textwidth]{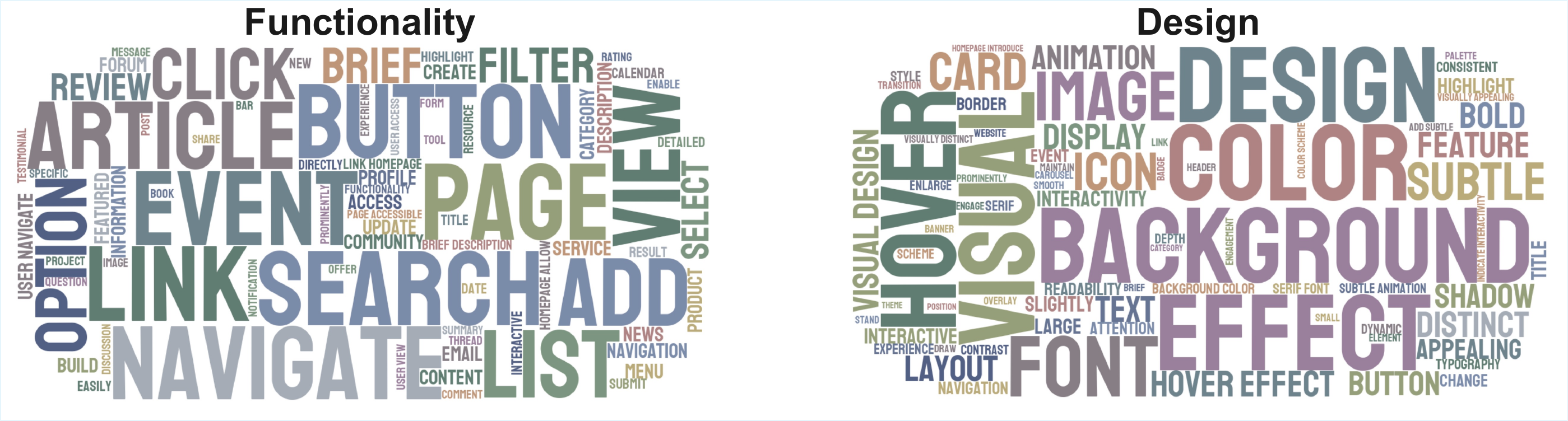}}
        \caption{Distributions of user intents for functionality and design types respectively.}
        \label{fig:data:intent_wordcloud}
\end{figure}

Wordcloud plots demonstrating distribution of user intents are in Figure \ref{fig:data:intent_wordcloud}.

A data example is as follows:
\begin{itemize}[leftmargin=*, itemsep=0pt, topsep=0pt]
    \item \textbf{Website goal:} The website "Shifting the Balance" appears to be a personal or family blog that shares thoughts and reflections on life, family, and societal issues.

\item \textbf{Intent (t=1):} \texttt{[type=functionality]} Build a website for "Shifting the Balance" that serves as a personal blog sharing thoughts on life, family, and societal issues. Implement a comment section accessible at the bottom of each blog post page, allowing users to leave comments and replies. Include an email subscription feature on the homepage and each blog post page, enabling visitors to subscribe with their email addresses to receive notifications about new posts. Provide social media sharing options for Twitter, Facebook, and LinkedIn on each blog post page to facilitate easy sharing of content. \begin{itemize}[leftmargin=*, itemsep=0pt, topsep=0pt]
\item \textbf{\textit{Test cases:}}
\item Navigate to the bottom of any blog post page to find the comment section. \texttt{[pass]} Users can see a text input field to write comment and an option to submit it. \texttt{[fail]} Users cannot write comment, or the comment cannot be submitted.
\item On the homepage and any blog post page, locate the email subscription feature. \texttt{[pass]} Users can see an input field to input email address, and an option to confirm subscription. \texttt{[fail]} The input field for email address is missing, or the email address cannot be submitted.
\item On any blog post page, find the social media sharing options. \texttt{[pass]} Users can see buttons or links for sharing on Twitter, Facebook, and LinkedIn. \texttt{[fail]} One or more of the social media sharing options is missing.
\end{itemize}
\item \textbf{Intent (t=2):} \texttt{[type=functionality]} Enhance the comment section by adding a "Like" feature to each comment and reply, allowing users to express appreciation for comments. Ensure that each comment and reply has a visible "Like" button next to it, which increments a visible like counter when clicked. Users should be able to navigate to the comment section by scrolling to the bottom of any blog post page from the homepage. \begin{itemize}[leftmargin=*, itemsep=0pt, topsep=0pt]
\item \textbf{\textit{Test cases:}}
\item Each comment and reply in the comment section of any blog post page has a visible "Like" button next to it. \texttt{[pass]} Every comment and reply displays a "Like" button adjacent to it. \texttt{[fail]} One or more comments or replies do not have a "Like" button next to them.
\item Clicking the "Like" button next to each comment or reply in the comment section at the bottom of any blog post page increases its like counter by one. \texttt{[pass]} The like counter for a comment or reply increases by one when its "Like" button is clicked. \texttt{[fail]} The like counter does not change, or changes incorrectly, when the "Like" button is clicked.
\item The like counter is displayed for each comment and reply in the comment section at the bottom of each blog post page. \texttt{[pass]} A like counter is visibly displayed for every comment and reply. \texttt{[fail]} The like counter is missing for one or more comments or replies.
\end{itemize}
\item \textbf{Intent (t=3):} \texttt{[type=functionality]} Introduce a "Featured Posts" section on the homepage, showcasing a rotating selection of three highlighted blog posts to draw visitor attention. Ensure that users can navigate to the detailed blog post page by clicking on any featured post title or image. Include a brief excerpt and a placeholder image for each featured post to provide a preview of the content. The rotation should automatically change the featured posts every 10 seconds, but users should also have the option to manually navigate through them using forward and backward arrows. \begin{itemize}[leftmargin=*, itemsep=0pt, topsep=0pt]
\item \textbf{\textit{Test cases:}}
\item The homepage displays a "Featured Posts" section with three distinct blog post previews. \texttt{[pass]} The homepage shows a section labeled "Featured Posts" with three previews of different blog posts. \texttt{[fail]} The "Featured Posts" section is missing, or it does not contain exactly three blog post previews.
\item Each featured post in the "Featured Posts" section on the homepage has a visible title, excerpt, and placeholder image. \texttt{[pass]} All three featured posts show a title, an excerpt, and a placeholder image. \texttt{[fail]} Any of the featured posts is missing a title, an excerpt, or a placeholder image.
\item Clicking on a featured post's title or image on the homepage navigates the user to the detailed blog post page. \texttt{[pass]} Clicking on the title or image of a featured post redirects the user to its detailed blog post page. \texttt{[fail]} Clicking on the title or image does not navigate to the detailed blog post page.
\item The featured posts on the homepage automatically rotate every 10 seconds. \texttt{[pass]} The set of featured posts changes automatically every 10 seconds. \texttt{[fail]} The featured posts do not change automatically, or the timing is incorrect.
\item Users can manually navigate through the featured posts on the homepage using forward and backward arrows. \texttt{[pass]} The user can click forward and backward arrows to manually navigate through the featured posts. \texttt{[fail]} The arrows are missing, or clicking them does not navigate through the featured posts.
\end{itemize}
\item \textbf{Intent (t=4):} \texttt{[type=design]} Refine the visual design of the blog post pages by introducing a cohesive color scheme and typography style that enhances readability and aesthetic appeal. Use a serif font for blog titles and a sans-serif font for body text to create a clear hierarchy and visual contrast. Apply a soft background color to the entire page to reduce eye strain and highlight the content areas. Ensure that interactive elements such as social media sharing buttons, comment section, and "Like" buttons have a distinct hover effect to visually indicate interactivity. \begin{itemize}[leftmargin=*, itemsep=0pt, topsep=0pt]
\item \textbf{\textit{Test cases:}}
\item Blog titles on each blog post page are displayed in a serif font, while body text uses a sans-serif font. \texttt{[pass]} Blog titles appear in a serif font, and body text is in a sans-serif font. \texttt{[fail]} Both blog titles and body text appear in the same font style.
\item The background color of each blog post page is soft and distinct from the content areas. \texttt{[pass]} The page background is a soft color, differing from the content sections. \texttt{[fail]} The page background color is harsh or indistinct from content areas.
\item Interactive elements such as social media sharing buttons, the comment section, and "Like" buttons on each blog post page exhibit a distinct visual change on hover. \texttt{[pass]} Interactive elements change visually (e.g., color, size) when hovered over. \texttt{[fail]} Interactive elements show no visual change when hovered over.
\end{itemize}
\item \textbf{Intent (t=5):} \texttt{[type=design]} Enhance the homepage by designing an inviting hero section at the top, featuring a large, captivating background image that represents the blog's themes. Overlay this image with a semi-transparent color layer to ensure text readability. Include a welcoming headline in a bold serif font, accompanied by a concise tagline in a smaller sans-serif font. Add a call-to-action button, styled with a distinct hover effect, encouraging visitors to explore the blog further. \begin{itemize}[leftmargin=*, itemsep=0pt, topsep=0pt]
\item \textbf{\textit{Test cases:}}
\item The hero section on the homepage features a large background image. \texttt{[pass]} A large background image is prominently displayed at the top of the homepage. \texttt{[fail]} There is no background image, or it is not prominently displayed.
\item The homepage's hero section should have a background image with a semi-transparent color overlay. \texttt{[pass]} The background image has a visible semi-transparent color layer over it. \texttt{[fail]} The background image does not have any overlay, or the overlay is not semi-transparent.
\item The headline is in a bold serif font, and the tagline is in a smaller sans-serif font. \texttt{[pass]} The headline uses a bold serif font, and the tagline uses a smaller sans-serif font. \texttt{[fail]} The headline and tagline do not follow the specified typography styles.
\item The call-to-action button on the homepage hero section is present with a distinct hover effect. \texttt{[pass]} A call-to-action button is visible and exhibits a distinct visual change when hovered over. \texttt{[fail]} There is no call-to-action button, or it lacks a hover effect.
\end{itemize}
\item \textbf{Intent (t=6):} \texttt{[type=design]} Enhance the visual appeal of the "Featured Posts" section by adding smooth transition animations when rotating between posts, creating a more engaging user experience. Introduce a subtle zoom-in effect on the placeholder images of each featured post when hovered over, emphasizing interactivity and drawing attention. Ensure the navigation arrows have a bold, contrasting color to make them easily identifiable, and implement a gentle bounce animation on hover to indicate their functionality. \begin{itemize}[leftmargin=*, itemsep=0pt, topsep=0pt]
\item \textbf{\textit{Test cases:}}
\item When the "Featured Posts" section on the homepage rotates between posts, there is a smooth transition animation observable. \texttt{[pass]} The transition between posts is smooth and visually fluid. \texttt{[fail]} The transition appears abrupt or jerky.
\item Hovering over a placeholder image in the "Featured Posts" section on the homepage results in a noticeable zoom-in effect. \texttt{[pass]} The image enlarges slightly when hovered over. \texttt{[fail]} The image remains static with no change on hover.
\item The navigation arrows in the "Featured Posts" section on the homepage are in a bold, contrasting color compared to the background. \texttt{[pass]} The arrows stand out distinctly against the background. \texttt{[fail]} The arrows blend into the background or are difficult to discern.
\item Hovering over the navigation arrows in the "Featured Posts" section on the homepage triggers a gentle bounce animation. \texttt{[pass]} The arrows exhibit a bounce effect when hovered over. \texttt{[fail]} The arrows show no movement or animation on hover.
\end{itemize}
\item \textbf{Intent (t=7):} \texttt{[type=functionality]} Introduce a search functionality on the homepage, allowing users to easily find specific blog posts by entering keywords. Ensure the search bar is prominently positioned near the top of the homepage and includes a placeholder text prompting users to "Search blog posts...". Upon entering a search query and hitting enter, users should be directed to a results page displaying a list of blog posts that match the search terms. Each result should include the blog post title, a brief excerpt, and a link to the full post for easy navigation. \begin{itemize}[leftmargin=*, itemsep=0pt, topsep=0pt]
\item \textbf{\textit{Test cases:}}
\item Verify the presence of a search bar near the top of the homepage with placeholder text "Search blog posts...". \texttt{[pass]} The search bar is visible near the top of the homepage with the specified placeholder text. \texttt{[fail]} The search bar is not visible, or the placeholder text is incorrect or missing.
\item Test the search functionality on the homepage by entering a keyword in the search bar and pressing enter. \texttt{[pass]} The user is directed to a results page displaying a list of blog posts matching the keyword. \texttt{[fail]} The user is not directed to a results page, or no relevant posts are displayed despite matching content.
\item Check that each search result on the results page includes the blog post title, a brief excerpt, and a link to the full post. \texttt{[pass]} Each result displays the title, excerpt, and a clickable link to the full post. \texttt{[fail]} Any of the required elements (title, excerpt, link) are missing from the search results.
\end{itemize}
\item \textbf{Intent (t=8):} \texttt{[type=functionality]} Introduce a "Related Posts" section at the end of each blog post page, featuring three blog posts with similar themes to engage readers further. Ensure that each related post displays a title, a brief excerpt, and a placeholder image, with links to navigate to the full blog post. Allow users to access this section by scrolling to the bottom of a blog post page, just above the comment section. Incorporate a feature that dynamically selects related posts based on shared tags or categories with the current post. \begin{itemize}[leftmargin=*, itemsep=0pt, topsep=0pt]
\item \textbf{\textit{Test cases:}}
\item Navigate to the bottom of each blog post page and verify the presence of a "Related Posts" section. \texttt{[pass]} The "Related Posts" section is visible just above the comment section. \texttt{[fail]} The "Related Posts" section is not visible or appears below the comment section.
\item Check that each related post in the "Related Posts" section at the end of each blog post page displays a title, a brief excerpt, and a placeholder image. \texttt{[pass]} Each related post shows a title, an excerpt, and an image. \texttt{[fail]} One or more related posts are missing a title, an excerpt, or an image.
\item For each post in the "Related Posts" section at the end of each blog post page, users can easily navigate to their respective full blog post. \texttt{[pass]} Users can navigate to the full blog post, such as by clicking the title, the image, or a clickable link. \texttt{[fail]} Clicking a related post title or image does not redirect to the corresponding full post page, and there are no visible navigation links or buttons.
\item Verify that the related posts on each blog post page are selected based on shared tags or categories with the current post. \texttt{[pass]} The related posts share at least one tag or category with the current post. \texttt{[fail]} None of the related posts share any tags or categories with the current post.
\end{itemize}
\item \textbf{Intent (t=9):} \texttt{[type=design]} Enhance the "Related Posts" section by incorporating a visually appealing card layout that neatly organizes each related post with a subtle drop shadow and rounded corners. Use a consistent background color across all cards to unify the section visually. Implement an interactive hover effect on each card, slightly elevating it to indicate interactivity and engagement. Ensure that the title of each related post is in a bold serif font for emphasis, while excerpts remain in a smaller sans-serif font for readability. \begin{itemize}[leftmargin=*, itemsep=0pt, topsep=0pt]
\item \textbf{\textit{Test cases:}}
\item Each related post is displayed within a card with rounded corners and a subtle drop shadow on the blog post page. \texttt{[pass]} Each related post appears within a bordered card with rounded corners and a shadow effect. \texttt{[fail]} Related posts lack rounded corners or a shadow effect, or do not appear within a card layout.
\item All cards in the "Related Posts" section on each blog post page have a consistent background color. \texttt{[pass]} The background color is uniform across all related post cards. \texttt{[fail]} The background color varies between related post cards.
\item A hover effect elevates each card in the "Related Posts" section at the end of each blog post page slightly when interacted with. \texttt{[pass]} Hovering over a related post card causes it to slightly elevate from the rest. \texttt{[fail]} Hovering over a related post card does not change its position or appearance.
\item The title of each related post in the "Related Posts" section at the end of each blog post page is in a bold serif font, while excerpts are in a smaller sans-serif font. \texttt{[pass]} Titles are distinctly bold and serif, and excerpts are visibly smaller and sans-serif. \texttt{[fail]} Titles and excerpts do not follow the specified font styling, lacking contrast in weight or style.
\end{itemize}
\item \textbf{Intent (t=10):} \texttt{[type=design]} Enhance the visual design of the email subscription feature across the homepage and each blog post page by introducing a stylish subscription form. Use a rounded input field with a placeholder text that invites users to "Enter your email...". Add a vibrant call-to-action button next to the input field, styled with a hover effect that changes its background color. Ensure the form is positioned prominently within the layout, and add a subtle animation that draws attention to the form upon page load. \begin{itemize}[leftmargin=*, itemsep=0pt, topsep=0pt]
\item \textbf{\textit{Test cases:}}
\item The email subscription form on the homepage and each blog post page should feature a rounded input field with placeholder text "Enter your email...". \texttt{[pass]} The input field is rounded and displays the specified placeholder text. \texttt{[fail]} The input field is not rounded or does not display the specified placeholder text.
\item The call-to-action button next to the input field in the email subscription form should have a vibrant background color and change color on hover. \texttt{[pass]} The button has a vibrant background color and visibly changes color when hovered over. \texttt{[fail]} The button does not have a vibrant background color or does not change color on hover.
\item The email subscription form should be prominently positioned within the layout of the homepage and each blog post page. \texttt{[pass]} The form is easily noticeable and positioned in a prominent area of the page. \texttt{[fail]} The form is not easily noticeable or is positioned in a less prominent area of the page.
\item The email subscription form on the homepage and each blog post page should feature a subtle animation upon page load to draw attention. \texttt{[pass]} A subtle animation is visible when the page loads, drawing attention to the form. \texttt{[fail]} No animation is present upon page load, or the animation does not draw attention to the form.
\end{itemize}

\end{itemize}

\appsection{Experimental Details}
\label{sec:appendix:exp_detail}

We employ vLLM \citep{kwon2023efficient} for model serving. We set the maximum output token limit empirically for each model family, aiming to  allow the model to complete generation for most ($>$95\%) input instructions.
Specifically, the limits are set as follows: 10,000 for GPT-4o, Llama-3.3, Qwen2.5-VL, GPT-OSS and Gemma3 models; 20,000 for Gemini-2.5-Pro, Qwen3, Qwen3-Coder, DeepSeek-R1, GLM-4, Ovis2.5, and Qwen3-VL models; and 30,000 for Claude-4-Sonnet. Models receive the full conversation history as input during inference. If the context limit is exceeded, we truncate the history by removing model responses from the earliest turns while always preserving the user instructions. 
Detailed \texttt{huggingface} ID for open-source models are in Table \ref{tab:hf_id}.
\begin{table}[!htb]
    \centering
    \begin{tabular}{ll|l}
    \hline
    Model & Size & HF ID \\
         \hline
Qwen3 & 8B & Qwen/Qwen3-8B \\
GPT-OSS & 20B & openai/gpt-oss-20b \\
Qwen3-Coder & 30B & Qwen/Qwen3-Coder-30B-A3B-Instruct \\
Llama-3.3 & 70B & meta-llama/Llama-3.3-70B-Instruct \\
GPT-OSS & 120B & openai/gpt-oss-120b \\
Qwen3 & 235B & Qwen/Qwen3-235B-A22B \\
Qwen3-Coder & 480B & Qwen/Qwen3-Coder-480B-A35B-Instruct-FP8 \\
DeepSeek-R1 & 685B & deepseek-ai/DeepSeek-R1-0528 \\
\hline
Qwen2.5-VL & 7B & Qwen/Qwen2.5-VL-7B-Instruct \\
GLM-4.1V-Thinking & 9B & zai-org/GLM-4.1V-9B-Thinking \\
Ovis2.5 & 9B & AIDC-AI/Ovis2.5-9B \\
Gemma3 & 12B & google/gemma-3-12b-it \\
Gemma3 & 27B & google/gemma-3-27b-it \\
Qwen3-VL & 30B & Qwen/Qwen3-VL-30B-A3B-Instruct \\
Qwen2.5-VL & 32B & Qwen/Qwen2.5-VL-32B-Instruct \\
Qwen2.5-VL & 72B & Qwen/Qwen2.5-VL-72B-Instruct \\
GLM-4.5V & 108B & zai-org/GLM-4.5V \\
    \hline
    \end{tabular}
    \caption{Detailed \texttt{huggingface} ID for open-source models we evaluate.}
    \label{tab:hf_id}
\end{table}

\appsection{Implementation Details for User Simulator}
\label{sec:appendix:user_simulator}

\subsection{Textual Instruction Simulator}

As described in \S\ref{sec:dataset:user_simulation}, the textual instruction simulator is an LLM prompted with the current code state and asked to rewrite the user intent to (1) resolve ambiguity and (2) remove redundancy. The prompts for rewriting functionality- and design-type user intents are provided in \autoref{prompt:textual_user_f} and \autoref{prompt:textual_user_d}, respectively. We prompt GPT-4o to simulate textual instructions. Example outputs for the user intents in \S\ref{sec:appendix:data_details} are in Table \ref{tab:textual_inst}.

\begin{table}[!b]
    \centering
    \setlength{\tabcolsep}{3pt}
    \small
\begin{tabular}{l|p{0.42\linewidth}|p{0.52\textwidth}}
\hline
         \textbf{t}&\textbf{User Intents}&\textbf{Simulated Instructions}  \\
         \hline
 2 &    Enhance the comment section by adding a "Like" feature to each comment and reply, allowing users to express appreciation for comments. Ensure that each comment and reply has a visible "Like" button next to it, which increments a visible like counter when clicked. Users should be able to navigate to the comment section by scrolling to the bottom of any blog post page from the homepage.    & To enhance user engagement in the comment section of each blog post, add a "Like" feature to both comments and replies. Implement a "Like" button next to each comment and reply, which should visibly increase the like counter upon being clicked. {\color{darkgold}Ensure this functionality is available within the existing comment section located at the bottom of each blog post page.} Users can access the comment section by scrolling down to the end of any blog post, {\color{darkgold}which they can navigate to from the homepage by selecting a blog post title under "Latest Blog Posts".} \\
 \hline
 4 & Refine the visual design of the blog post pages by introducing a cohesive color scheme and typography style that enhances readability and aesthetic appeal. Use a serif font for blog titles and a sans-serif font for body text to create a clear hierarchy and visual contrast. Apply a soft background color to the entire page to reduce eye strain and highlight the content areas. Ensure that interactive elements such as social media sharing buttons, comment section, and "Like" buttons have a distinct hover effect to visually indicate interactivity. & Refine the visual design of the blog post pages by implementing a cohesive color scheme and typography style to enhance readability and aesthetic appeal. Introduce a serif font for blog titles and a sans-serif font for body text to establish a clear hierarchy and visual contrast. Apply a soft background color to the entire page to reduce eye strain and highlight the content areas. Ensure that interactive elements, such as the social media sharing buttons within each blog post, the comment section, and the "Like" buttons adjacent to comments, have a distinct hover effect to visually indicate interactivity. {\color{darkgold}Note that if the distinct hover effect is not currently visible, it must be implemented to enhance user experience.} \\
 \hline
 8 & Introduce a "Related Posts" section at the end of each blog post page, featuring three blog posts with similar themes to engage readers further. Ensure that each related post displays a title, a brief excerpt, and a placeholder image, with links to navigate to the full blog post. Allow users to access this section by scrolling to the bottom of a blog post page, just above the comment section. Incorporate a feature that dynamically selects related posts based on shared tags or categories with the current post.& To enhance user engagement on the blog post pages, introduce a "Related Posts" section directly above the existing comments section on each blog post page. This section should display three posts {\color{darkgold}that share similar themes with the current post, determined by matching tags or categories}. Each related post should include a title, a short excerpt, and a placeholder image, all of which should be clickable links that take users to the full blog post. This feature will help keep readers interested and encourage them to explore more content on the site. Ensure this section is seamlessly integrated so that users can easily find it by scrolling to the end of the blog post content.\\
 \hline
 10 & Enhance the visual design of the email subscription feature across the homepage and each blog post page by introducing a stylish subscription form. Use a rounded input field with a placeholder text that invites users to "Enter your email...". Add a vibrant call-to-action button next to the input field, styled with a hover effect that changes its background color. Ensure the form is positioned prominently within the layout, and add a subtle animation that draws attention to the form upon page load. & Refine the email subscription feature on the homepage and each blog post page by enhancing the subscription form's design. Implement a rounded input field with the placeholder text "Enter your email..." and position a vibrant call-to-action button next to it. This button should have a hover effect that changes its background color to make it more engaging. Ensure the form is prominently placed within the layout to capture attention and add a subtle animation that highlights the form when the page loads. {\color{darkgold}Since the current form is basic, these enhancements are not yet implemented and must be addressed.} \\
         \hline
    \end{tabular}
    \caption{Example outputs for simulated textual instructions. Content in the instructions that are distinct from original user intents are highlighted in {\color{darkgold}gold}.}
    \label{tab:textual_inst}
\end{table}

\subsection{Visual Instruction Simulator}

For the visual instruction simulator, we equip it with drawing tools based on the \texttt{matplotlib} package. In the first turn, the model is provided with screenshots of every page on the website, as well as the coordinates of all visible elements. These coordinates can be used in later turns to guide annotations. At each turn, we provide a VLM (GPT-4o in this work) with the following tools:
\begin{enumerate}[leftmargin=*, itemsep=0pt, topsep=0pt]
\item Subplot organization: Tools to create and manage multiple subplots using \texttt{matplotlib}'s subplot functionality.
\item Layout visualization: Renders a layout sketch, represented by minimal HTML code, into an image. This process generates additional artifacts: the coordinates of every element on the new layout page. These coordinates are provided to the model in the next turn to assist with drawing.
\item Shape drawing: Tools for drawing various shapes using \texttt{matplotlib}.
\item Text annotation: A custom text annotation tool. We constrain the VLM to call our tool instead of \texttt{matplotlib}'s native text function. Our tool automatically organizes text annotations to minimize overlapping.
\end{enumerate}
The VLM will generate python scripts to draw a visual instruction image using these tools. If the generated code has errors during execution, we allow models to re-generate for up to 5 times per turn.
At the end of each tool-calling turn, the results, including the drawn images and other artifacts (e.g., coordinates), are fed back to the model. The model is then asked if the visualization is finished, i.e., if the visual instruction fully represents the user's intent. If not, it is allowed to perform another round of tool calling. We allow a maximum of three turns. The full prompt is in \autoref{prompt:visual_user}.

\begin{figure}[!h]
    \centering
    \includegraphics[width=0.8\linewidth]{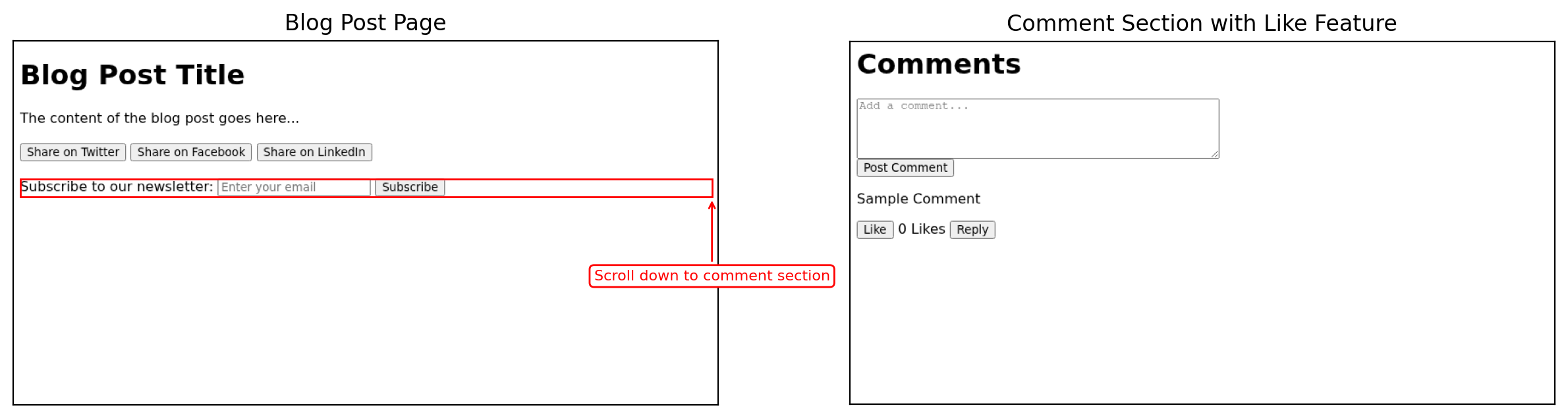}
    \caption{Intermediate drawing results after executing code in \autoref{code:visual_inst.1.0}.}
    \label{fig:visual_inst.1.0}
\end{figure}

\begin{figure}[!h]
    \centering
    \includegraphics[width=0.8\linewidth]{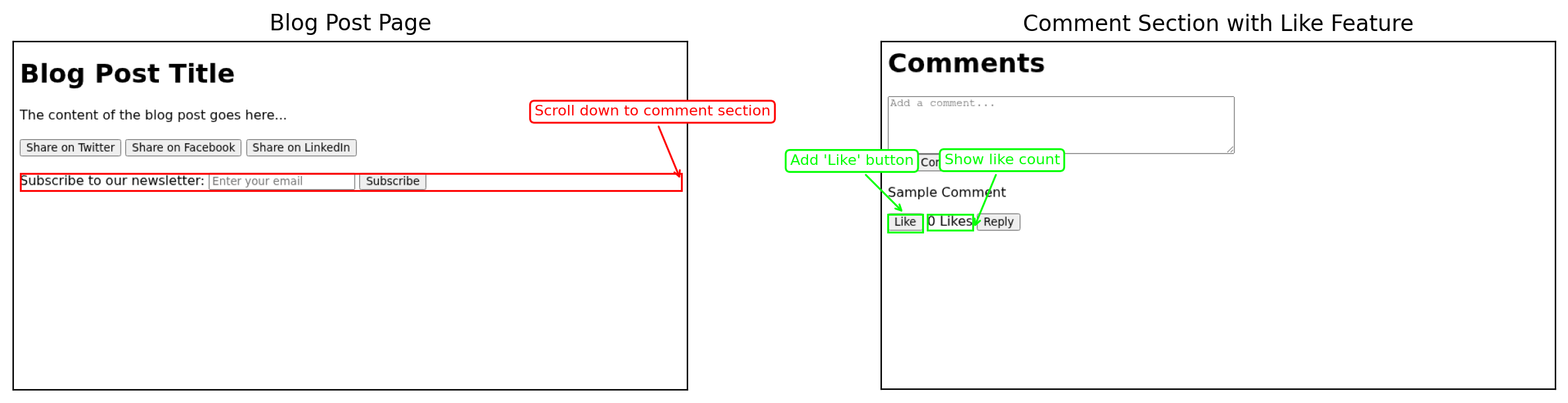}
    \caption{\small Simulated visual instructions for \texttt{t=2}. This is obtained after executing code in \autoref{code:visual_inst.1.1} on top of Figure \ref{fig:visual_inst.1.0}. User intent: Enhance the comment section by adding a "Like" feature to each comment and reply, allowing users to express appreciation for comments. Ensure that each comment and reply has a visible "Like" button next to it, which increments a visible like counter when clicked. Users should be able to navigate to the comment section by scrolling to the bottom of any blog post page from the homepage.}
    \label{fig:visual_inst.1}
\end{figure}

\begin{figure}[!p]
    \centering
    \includegraphics[width=0.8\linewidth]{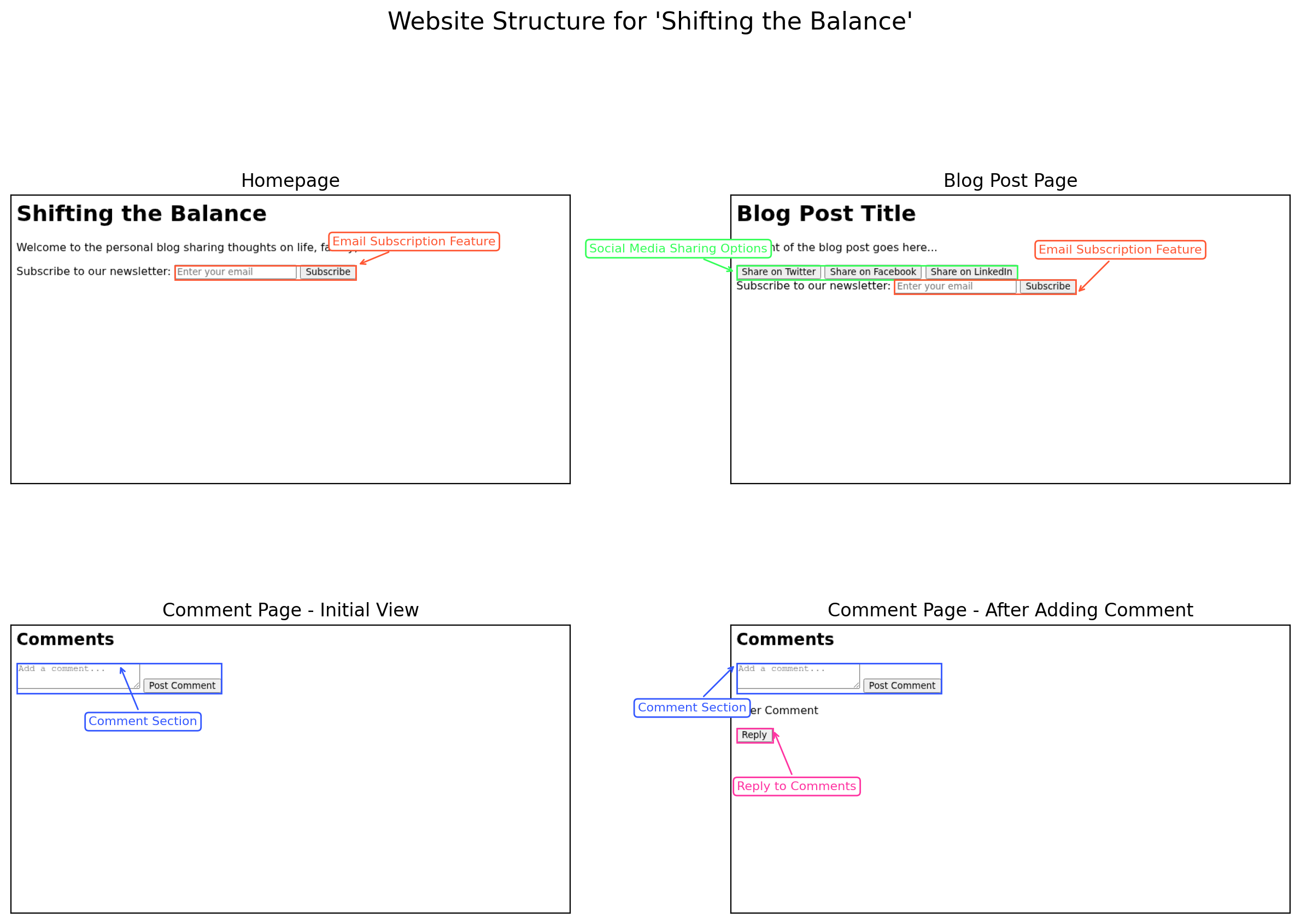}
    \caption{\small Simulated visual instructions for \texttt{t=1}. User intent: Build a website for "Shifting the Balance" that serves as a personal blog sharing thoughts on life, family, and societal issues. Implement a comment section accessible at the bottom of each blog post page, allowing users to leave comments and replies. Include an email subscription feature on the homepage and each blog post page, enabling visitors to subscribe with their email addresses to receive notifications about new posts. Provide social media sharing options for Twitter, Facebook, and LinkedIn on each blog post page to facilitate easy sharing of content.}
    \label{fig:visual_inst.0}
\end{figure}

\begin{figure}[!p]
    \centering
    \includegraphics[width=0.8\linewidth]{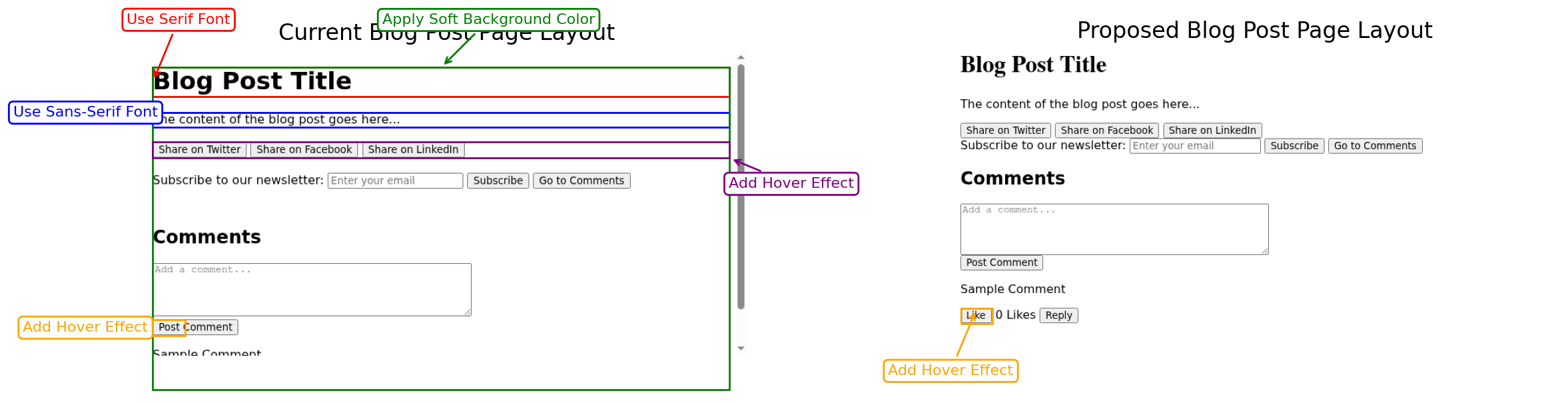}
    \caption{\small Simulated visual instructions for \texttt{t=4}. User intent: Refine the visual design of the blog post pages by introducing a cohesive color scheme and typography style that enhances readability and aesthetic appeal. Use a serif font for blog titles and a sans-serif font for body text to create a clear hierarchy and visual contrast. Apply a soft background color to the entire page to reduce eye strain and highlight the content areas. Ensure that interactive elements such as social media sharing buttons, comment section, and "Like" buttons have a distinct hover effect to visually indicate interactivity.}
    \label{fig:visual_inst.3}
\end{figure}

\begin{figure}[!h]
    \centering
    \includegraphics[width=0.8\linewidth]{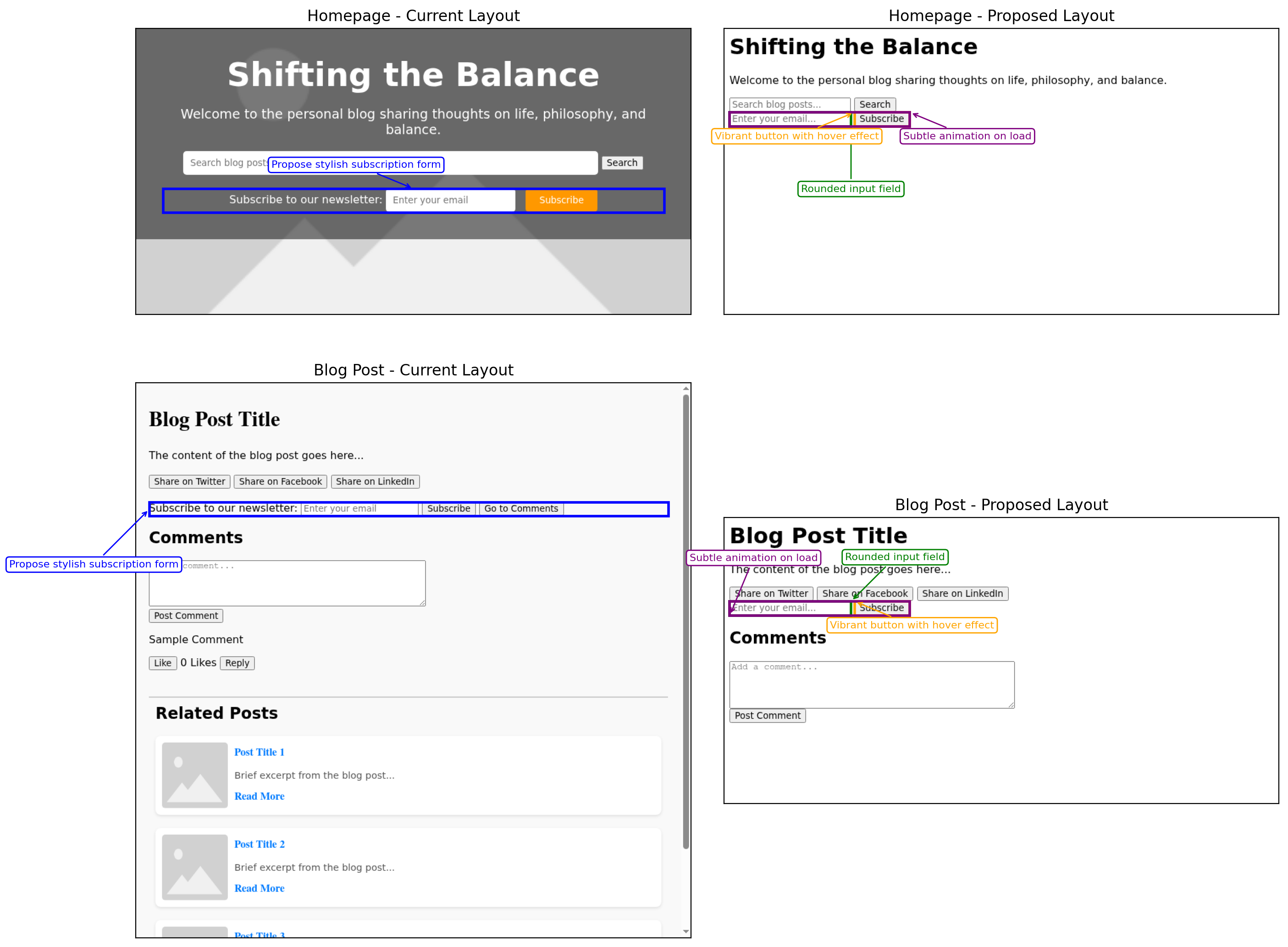}
    \caption{\small Simulated visual instructions for \texttt{t=10}. User intent: Enhance the visual design of the email subscription feature across the homepage and each blog post page by introducing a stylish subscription form. Use a rounded input field with a placeholder text that invites users to "Enter your email...". Add a vibrant call-to-action button next to the input field, styled with a hover effect that changes its background color. Ensure the form is positioned prominently within the layout, and add a subtle animation that draws attention to the form upon page load.}
    \label{fig:visual_inst.9}
\end{figure}

Example outputs for the user intents from \S\ref{sec:appendix:data_details} are shown in Figure \ref{fig:visual_inst.1}-\ref{fig:visual_inst.9}. To demonstrate the multi-turn drawing process, we show the code snippets in two turns of instruction drawing in \autoref{code:visual_inst.1.0} and \autoref{code:visual_inst.1.1}, with their corresponding output images in Figure \ref{fig:visual_inst.1.0} and \ref{fig:visual_inst.1}.

\begin{lstlisting}[basicstyle=\scriptsize\ttfamily, caption={\textbf{Code in the \textit{first} step in the visual instruction simulation process.} The resulted image is in Figure \ref{fig:visual_inst.1.0}. Our custom tools are highlighted in {\color{darkgold}gold}.}, label={code:visual_inst.1.0}]]
import matplotlib.pyplot as plt
from PIL import Image
import matplotlib.patches as patches

# Define the number of subplots
fig, axes = plt.subplots(1, 2, figsize=(12, 6))

# First Subplot: Blog Post Page
axes[0].imshow(Image.open("screenshot_blog-post.png"))
scroll_shape = patches.Rectangle((8, 160.9), 809, 21, facecolor='none', edgecolor='#FF0000')
axes[0].add_patch(scroll_shape)
(*@\textcolor{darkgold}{text\_annotation}@*)(axes[0], "Scroll down to comment section", scroll_shape, color='#FF0000')
axes[0].set_title("Blog Post Page")

# Second Subplot: Comment Section with Like Feature
html_code_comments = """
<html lang="en"><head>
    <meta charset="UTF-8">
    <meta name="viewport" content="width=device-width, initial-scale=1.0">
    <title>Comments</title>
    <style>
        body { font-family: Arial, sans-serif; }
        .comment-section { margin-top: 20px; }
        .comment { margin-top: 10px; }
    </style>
</head>
<body>
    <h1>Comments</h1>

    <div class="comment-section">
        <textarea id="comment" placeholder="Add a comment..." rows="4" cols="50"></textarea><br>
        <button onclick="postComment()">Post Comment</button>

        <div id="comments-list" class="comments-list">
            <div class="comment">
                <p>Sample Comment</p>
                <button onclick="likeComment(this)">Like</button> <span>0 Likes</span>
                <button onclick="replyToComment(this)">Reply</button>
            </div>
        </div>
    </div>
</body></html>
"""
(*@\textcolor{darkgold}{layout\_visualization}@*)(html_code_comments, axes[1])
axes[1].set_title("Comment Section with Like Feature")

plt.tight_layout()
\end{lstlisting}

\begin{lstlisting}[basicstyle=\scriptsize\ttfamily, caption={\textbf{Code in the \textit{second} step in the visual instruction simulation process.} The resulted image is in Figure \ref{fig:visual_inst.1}. Our custom tools are highlighted in {\color{darkgold}gold}.}, label={code:visual_inst.1.1}]]
# Adding annotations for the "Like" feature in the comment section

# Coordinates for the "Like" button and "0 Likes" counter
like_button_coords = (8, 209.4, 42.6, 21)
like_counter_coords = (55.7, 209.4, 55.5, 19)

# Drawing a rectangle around the "Like" button
like_button_shape = patches.Rectangle(
    (like_button_coords[0], like_button_coords[1]),
    like_button_coords[2],
    like_button_coords[3],
    facecolor='none',
    edgecolor='#00FF00'
)
axes[1].add_patch(like_button_shape)
(*@\textcolor{darkgold}{text\_annotation}@*)(axes[1], "Add 'Like' button", like_button_shape, color='#00FF00')

# Drawing a rectangle around the "0 Likes" counter
like_counter_shape = patches.Rectangle(
    (like_counter_coords[0], like_counter_coords[1]),
    like_counter_coords[2],
    like_counter_coords[3],
    facecolor='none',
    edgecolor='#00FF00'
)
axes[1].add_patch(like_counter_shape)
(*@\textcolor{darkgold}{text\_annotation}@*)(axes[1], "Show like count", like_counter_shape, color='#00FF00')
\end{lstlisting}

\appsection{Implementation Details for Agent-Based Evaluation}
\label{sec:appendix:agent}

\begin{figure}[!htbp]
    \centering
    \includegraphics[width=0.8\linewidth]{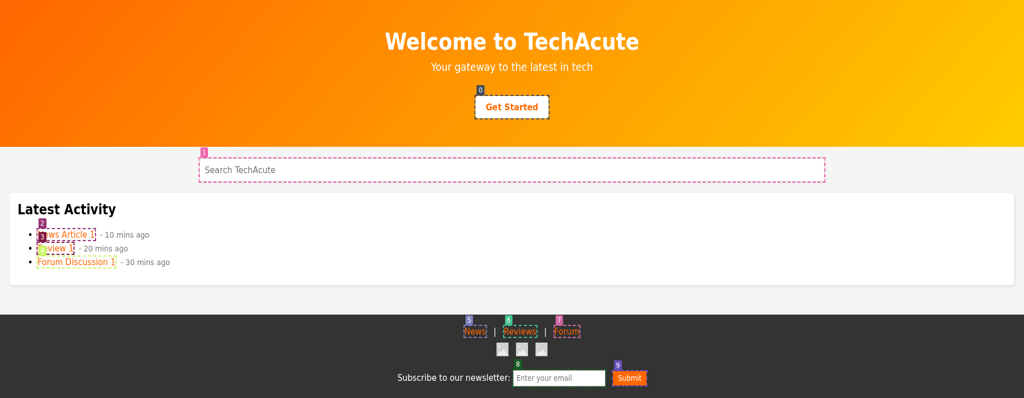}
    \caption{An example input observation for our evaluation agent. The textual representation is as follows: [0]: "Get Started"; [1]: <input> ""; [2]: "News Article 1"; [3]: "Review 1"; [4]: "Forum Discussion 1"; [5]: "News"; [6]: "Reviews"; [7]: "Forum"; [8]: <input> ""}
    \label{fig:example_obs}
\end{figure}

Following the WebVoyager framework \citep{he2024webvoyager}, our evaluation agent interacts with a web environment through multiple steps. At each step, the agent receives two forms of input observations, as shown in Figure \ref{fig:example_obs}: (1) A screenshot of the web page, annotated with numerical labels on all interactive elements, and (2) A corresponding text representation listing each labeled element and its properties.
Based on these observations, the agent selects an action from a predefined action space: \texttt{Click}, \texttt{Hover}, \texttt{Type}, \texttt{Select}, \texttt{Scroll}, \texttt{GoBack}, and \texttt{Upload}. Most actions require a \textit{target} argument, typically a numbered interactive element or the entire screen. Actions \textsc{Type}, \textsc{Select}, and \texttt{Upload} require an additional content argument, while \textsc{GoBack} requires no arguments.

\begin{figure}[!htbp]
    \centering
    \includegraphics[width=\linewidth]{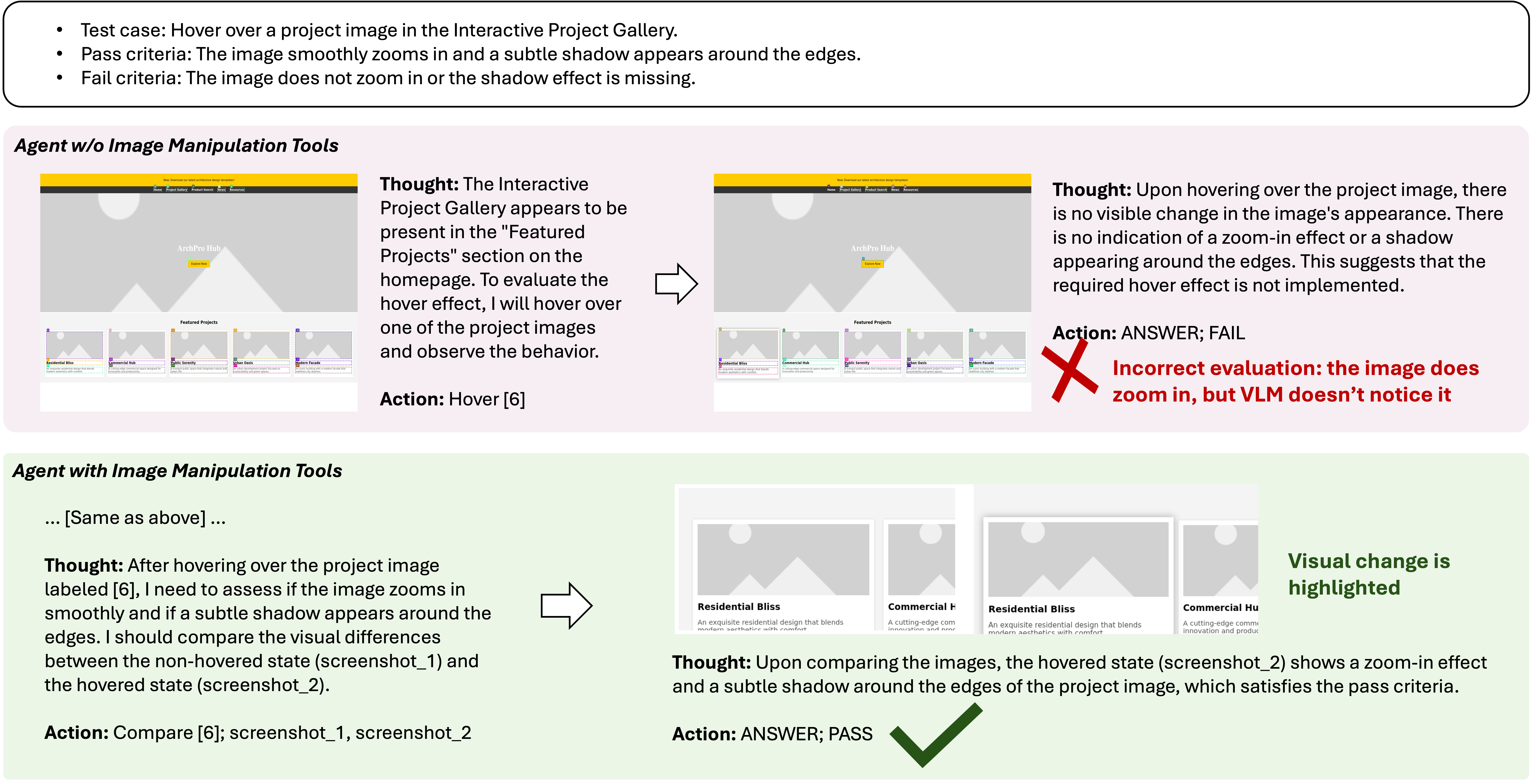}
    \caption{Example where evaluator without image manipulation tools fails, but evaluator with image manipulation tools succeeds.}
    \label{fig:imtool}
\end{figure}

We additionally augment the agent with \textbf{image manipulation tools}. These tools do not interact directly with the web environment, but instead operate on screenshots from prior turns. They are designed to sharpen the agent's perception and assist in judging test cases that require fine-grained visual perception, such as verifying a subtle hover effect in a local region. The available tools are:
\begin{enumerate}[leftmargin=*, itemsep=0pt, topsep=0pt]
    \item \texttt{ViewRaw screenshot\_x}: view the high-fidelity screenshot of a specific step, without annotated set-of-marks that may potentially hinder evaluation;
    \item \texttt{Compare [Numerical\_Label]; screenshot\_x, screenshot\_y}: Compare a specific element across two different steps. This tool will crop the local regions for specified element [Numerical\_Label] and concatenate the two crops across two screenshots into a single image for comparison.
    \item \texttt{ViewAnimation [Numerical\_Label or WINDOW]; screenshot\_x}: View the sequence of screenshots during the transition to \texttt{screenshot\_x} (from \texttt{screenshot\_x-1}). Similar to Compare, this tool concatenate the images into a single large image for comparison.
\end{enumerate}
Note that we use the \textit{raw and high-resolution} screenshots in  these tools, rather than the annotated and cluttered ones used for input observation.
The full prompt for our augmented agent is in \autoref{prompt:eval_pass_rate}. To show the benefits of our proposed image manipulation tools, Figure \ref{fig:imtool} shows an example where evaluator without image manipulation tools fails, but evaluator with image manipulation tools succeeds.

\appsection{Implementation Details for \method{}}

Detailed pseudo-code for \method{} is presented in Alg. \ref{alg:ours}. In the algorithm, the feedback function $\textsc{AgentFB}(\mathcal{M},\, \inst{},\, o) \to pass, expl$ queries a web agent powered by model $\mathcal{M}$ to verify whether instruction $\inst{}$ is well implemented in output $o$, and provides a binary feedback $pass$ along with a text explanation $expl$.

\renewcommand{\algorithmiccomment}[1]{\hfill\textcolor{LinkBlue}{\(\triangleright\) #1}}

\begin{algorithm}[!htbp]
\caption{\textbf{\method{} algorithm at each turn.}}
\label{alg:ours}
\begin{algorithmic}[1]
\Require Model $\mathcal{M}$;~~current-turn instruction $\inst{}_t$;~~history $\mathbf{H}_t$
\Require Feedback function $\textsc{AgentFB}(\mathcal{M}, \inst{}, o) \to pass, expl$
\Ensure Website $w$
\State $o_t \gets \mathcal{M}(H_{t-1}, x_t)$ \Comment{Draft output for turn $t$}
\State $F \gets \emptyset$ \Comment{Initialize feedback set}
\For{$j \gets 1$ \textbf{to} $t$}
    \State $pass_j, expl_j \gets \textsc{AgentFB}\left(\mathcal{M}, \inst{}_j, o_t\right)$
    \If{~$pass_j = \textbf{false}$~}
        \State $F \gets F \cup \{expl_j\}$ \Comment{Collect textual feedback}
    \EndIf
\EndFor
\If{$F \neq \emptyset$}
    \State $o_t \gets \mathcal{M}(\mathbf{H}_t, \inst{}_t, F)$ \Comment{Re-generate with feedback}
\EndIf
\State $w \gets \textsc{Render}(o_t)$
\State \Return $w$
\end{algorithmic}
\end{algorithm}

An example from our test set demonstrating \method{} is as follows. The simulated user instruction is -- Enhance the ``Local Attractions" page by adding a feature for users to create a personalized itinerary. This functionality will allow users to select attractions directly from the detailed view of each attraction and add them to their itinerary. Users should be able to include personal notes or preferences for each attraction they add. The itinerary should be accessible from the homepage, with options to view and print it. Ensure smooth navigation between the homepage and the hotel comparison page, allowing users to return without losing their itinerary details. Note that the itinerary feature does not currently exist, and thus must be implemented before these instructions can be fully applied.

A common implementation pitfall for this instruction is that an ``Add to Itinerary'' button is added to the interface, but is not functional. For the first iteration of model output $\mathcal{M}(H_{t-1}, x_t)$ exhibiting this pitfall, the agent feedback $expl_t$ generated by \method{} is as follows:
\begin{lstlisting}[float=!h, basicstyle=\scriptsize\ttfamily]
Summary:
1. Started on the homepage and navigated to the "Local Attractions" page.
2. Attempted to add attractions to a personalized itinerary using the "Add to Itinerary" buttons, but these actions were not executable.
3. Accessed the "My Itinerary" page, which displayed an empty itinerary with only a "Print Itinerary" button, and no options to add notes or preferences.
4. Returned to the homepage to verify if itinerary details persist across navigation, but there was no indication of any preserved itinerary details.

Analysis:
- The website does not currently support the full functionality of creating a personalized itinerary as instructed.
- The "Add to Itinerary" buttons do not function, and there is no way to add personal notes or preferences.
- The itinerary is not populated with selected attractions, and there is no evidence of itinerary persistence when navigating between pages.

Missing Elements:
- A functional "Add to Itinerary" feature that allows for adding attractions with personal notes.
- Persistent itinerary details that remain accessible across different pages.
- A confirmation or feedback mechanism to indicate successful addition of attractions to the itinerary.
\end{lstlisting}

\begin{figure}[!p]
    \centering
    \includegraphics[width=\linewidth]{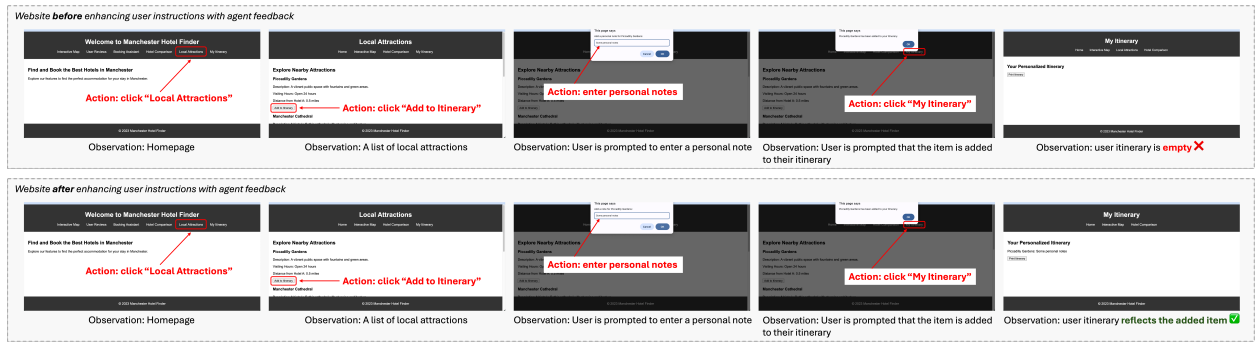}
    \caption{Screenshots of an example where \method{} successfully addresses a pitfall in the implementation.}
    \label{fig:ours_example}
\end{figure}

With the user instruction enhanced by this agent feedback, the model successfully addresses this issue and generates functional ``Add to Itinerary'' buttons. Detailed screenshots illustrating the website before and after using the agent feedback is shown in Figure \ref{fig:ours_example}.

\appsection{Error Cases}

\begin{figure}[!p]
    \centering
    \includegraphics[width=\linewidth]{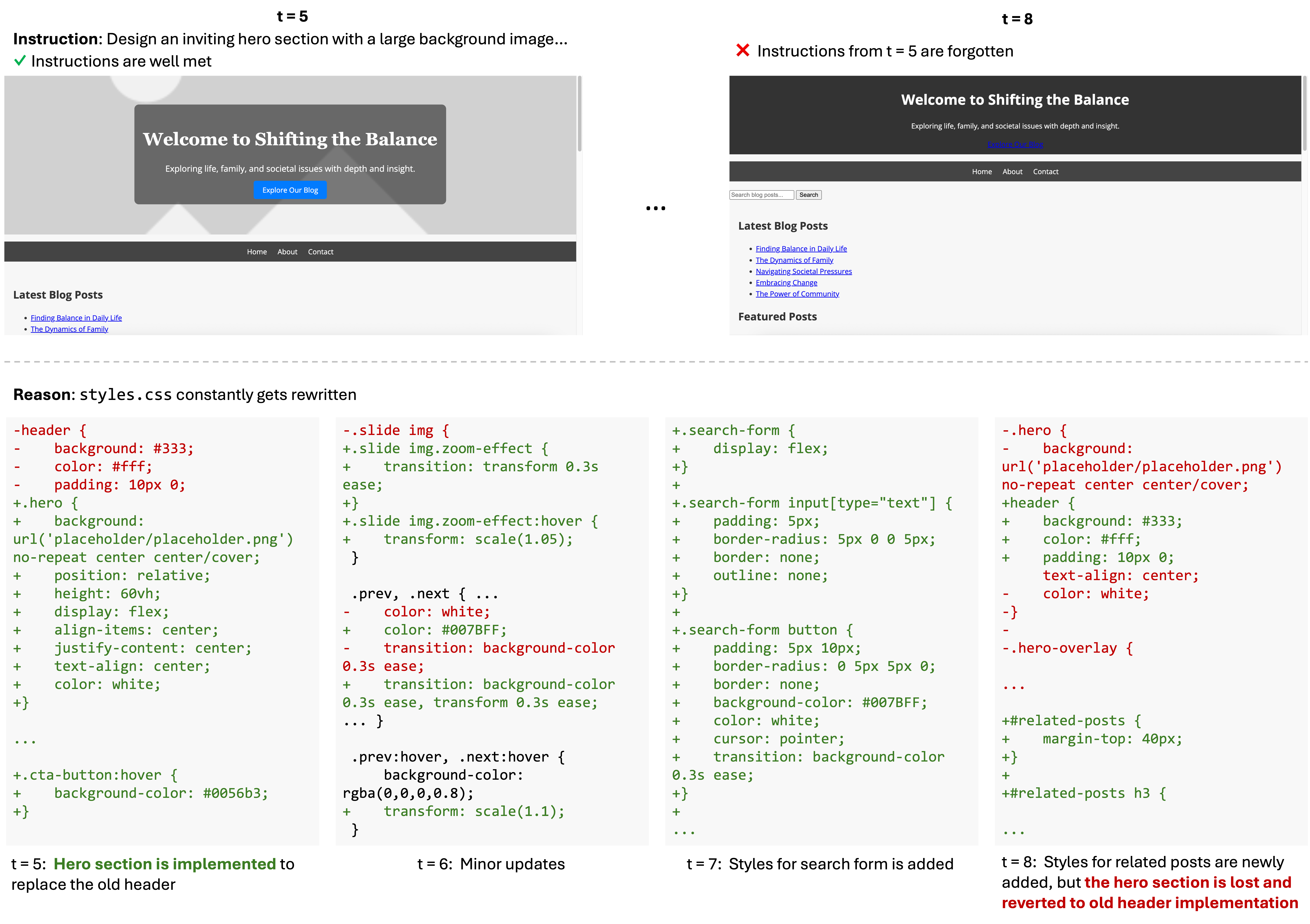}
    \caption{An example for forgetting issue. The hero section is correctly implemented in \texttt{t=5}, but gets forgotten at \texttt{t=8}. This is because the \texttt{styles.css} file constantly gets rewritten; at \texttt{t=8}, the rewriting lost the implementation of hero section, but reverts back to the old header implementation.}
    \label{fig:forgetting_example}
\end{figure}

\begin{figure}[!p]
    \centering
    \includegraphics[width=\linewidth]{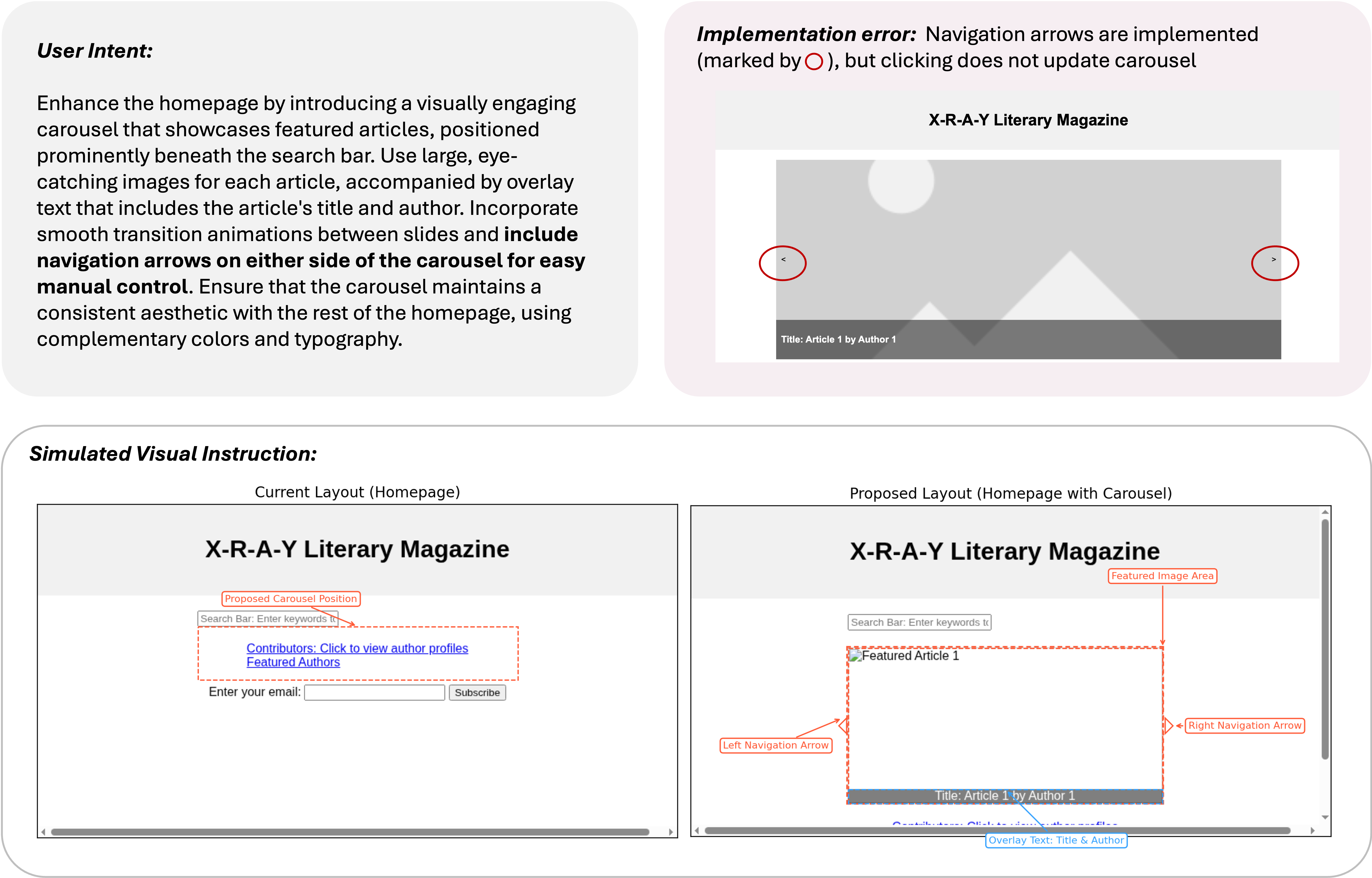}
    \caption{Error case for failure to implement implicit functionalities.}
    \label{fig:error1}
\end{figure}

\begin{figure}[!p]
    \centering
    \includegraphics[width=\linewidth]{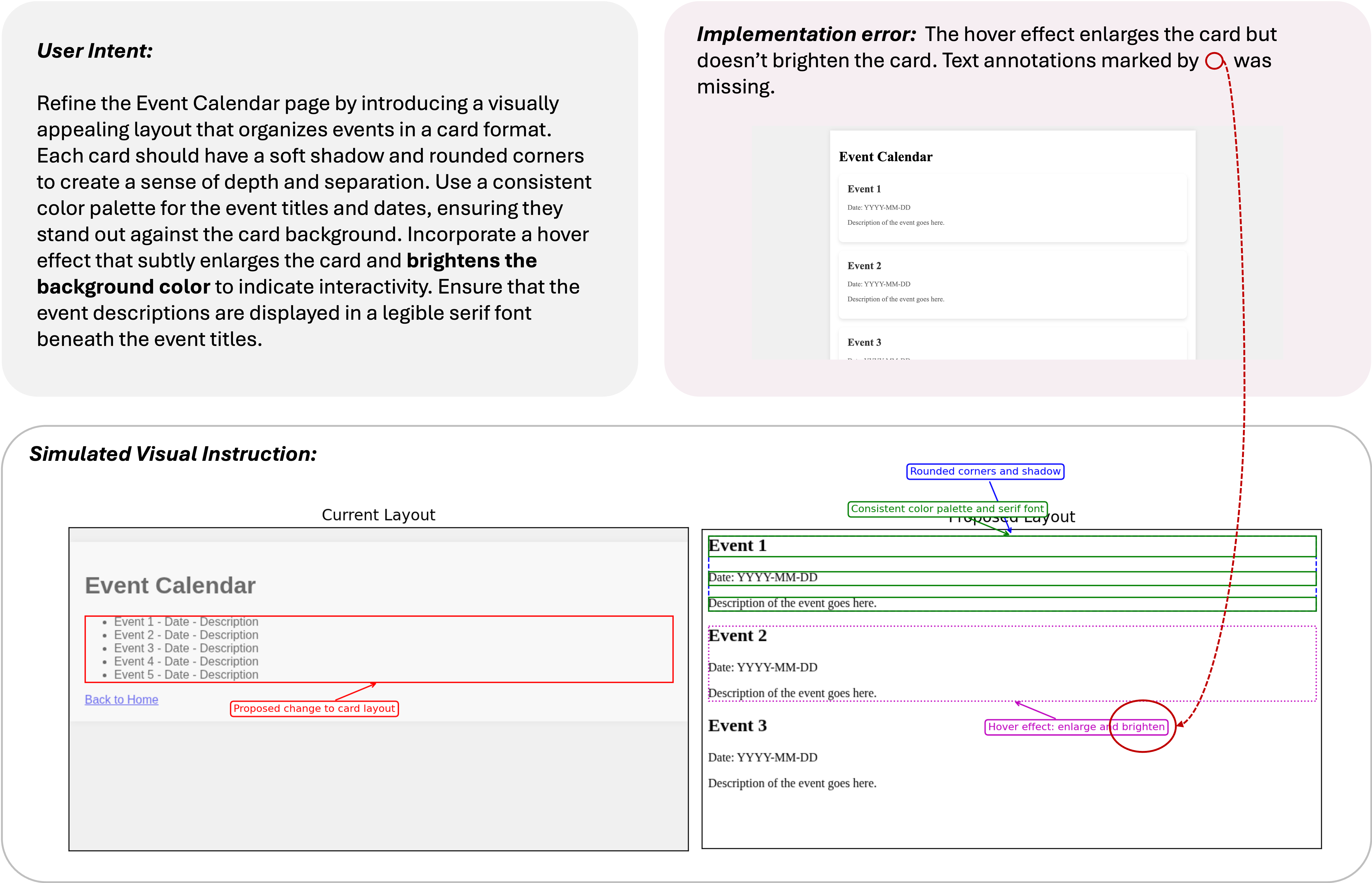}
    \caption{Error case for missing text
annotations.}
    \label{fig:error2}
\end{figure}

\begin{figure}[!t]
    \centering
    \includegraphics[width=\linewidth]{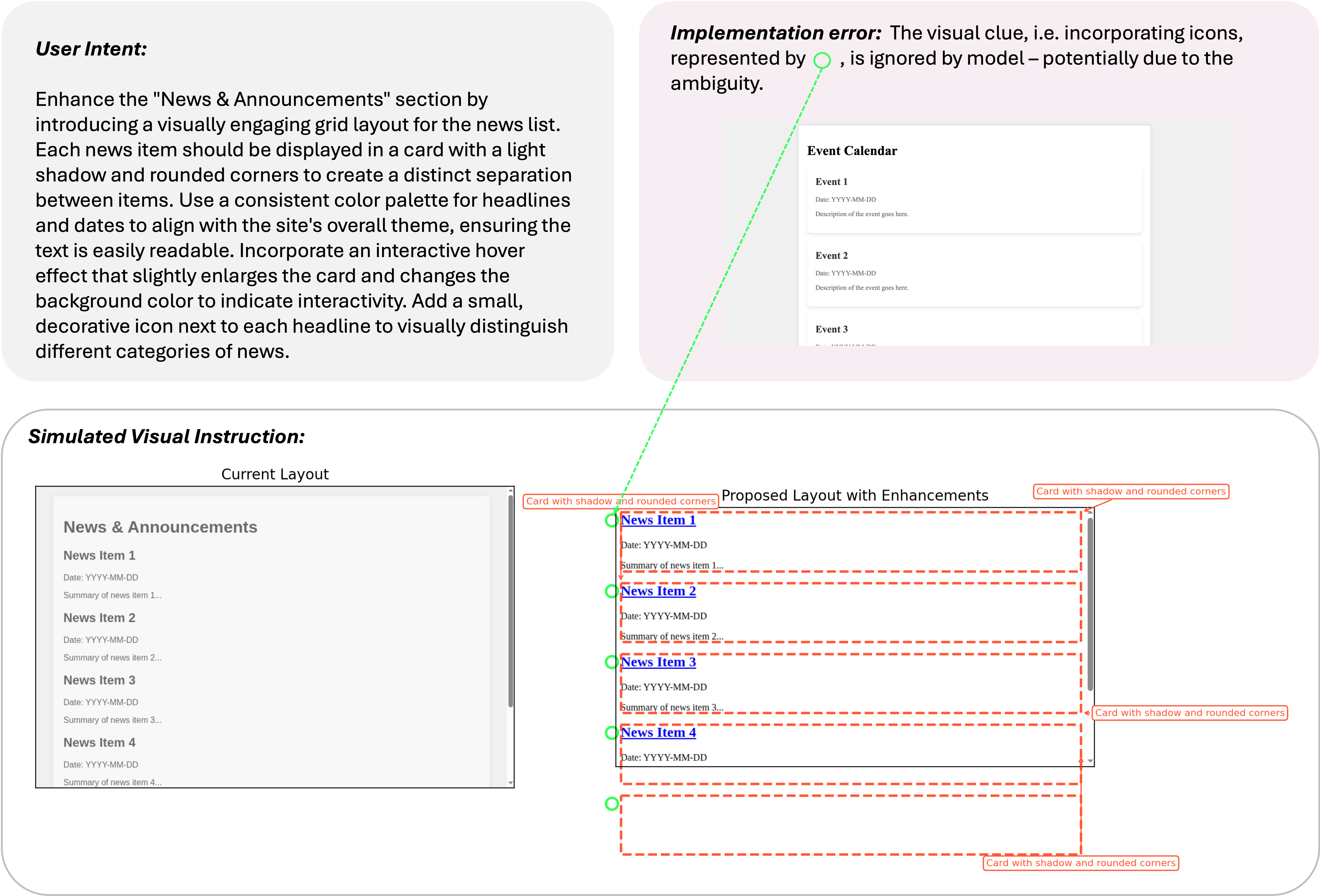}
    \caption{Error case for misinterpretation of visual clues.}
    \label{fig:error3}
\end{figure}

An example for forgetting issue is shown in Figure \ref{fig:forgetting_example}. Additionally, we show three error cases for the three types of visual instruction interpretation errors in Figure \ref{fig:error1}, \ref{fig:error2} and \ref{fig:error3} respectively.

\appsection{Example Trajectory from Usability Evaluation}

\begin{longtable}[!h]{>{\centering\arraybackslash}m{0.08\textwidth} m{0.25\textwidth} m{0.57\textwidth}}

    \caption{Example trajectory from usability evaluation.}
    \label{tab:my_long_table} \\
    
    \toprule
    \textbf{t} & \textbf{Screenshot} & \textbf{Thought and Action} \\
    \midrule
    \endfirsthead %
    
    \multicolumn{3}{l}{\tablename\ \thetable\ -- \textit{Continued from previous page}} \\
    \toprule
    \textbf{t} & \textbf{Screenshot} & \textbf{Thought and Action} \\
    \midrule
    \endhead %
    
    \midrule
    \multicolumn{3}{r}{\textit{Continued on next page}} \\
    \endfoot %
    
    \bottomrule
    \endlastfoot %
    
    1 %
    & %
      \includegraphics[width=\linewidth]{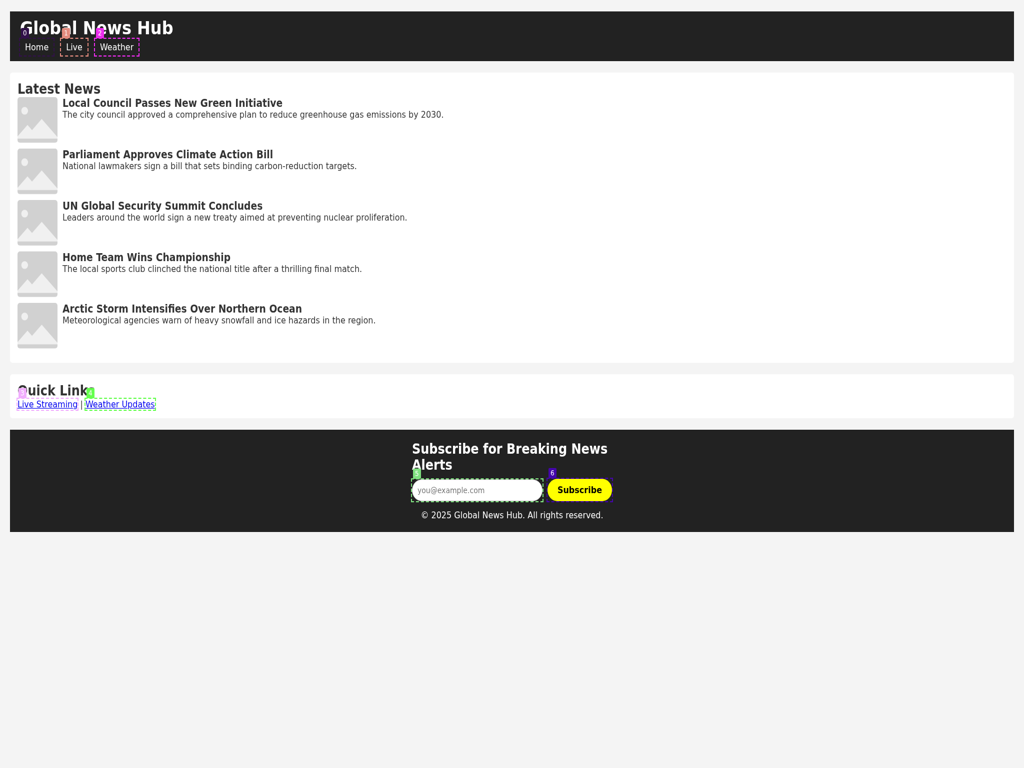}
    & \textbf{Thought:} The website appears to provide news articles with headlines related to local and international events. There are navigation options at the top for "Home," "Live," and "Weather," which likely lead to different sections of the site. There is also a subscription option for breaking news alerts. To explore more about the news coverage, I will click on "Live" to see if it provides real-time updates or live streaming news. \textbf{Action:} Click [1] \\ \midrule

    2 %
    & %
       \includegraphics[width=\linewidth]{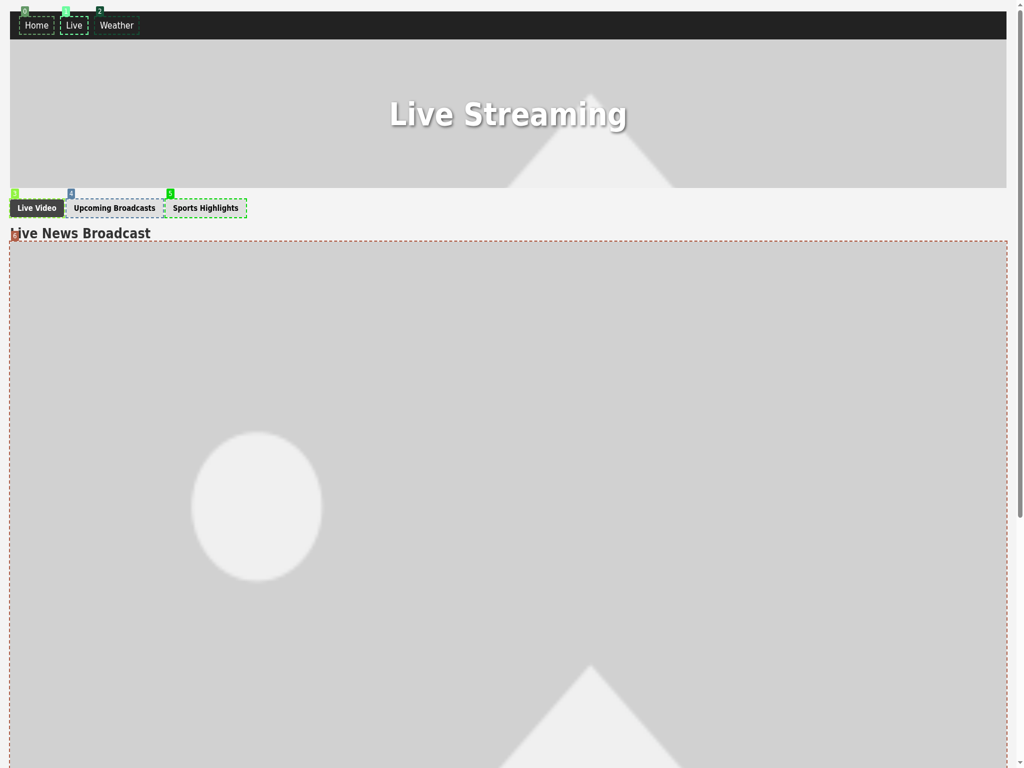} 
    & %
\textbf{Thought:} The "Live" section of the website offers options for live video, upcoming broadcasts, and sports highlights. This suggests that users can watch live news broadcasts or check schedules for future events. Since the goal is to understand how to access live news, I will click on "Live Video" to see if it provides access to a live news feed. \textbf{Action:} Click [3]
    \\ \midrule

    3 %
    & %
   \includegraphics[width=\linewidth]{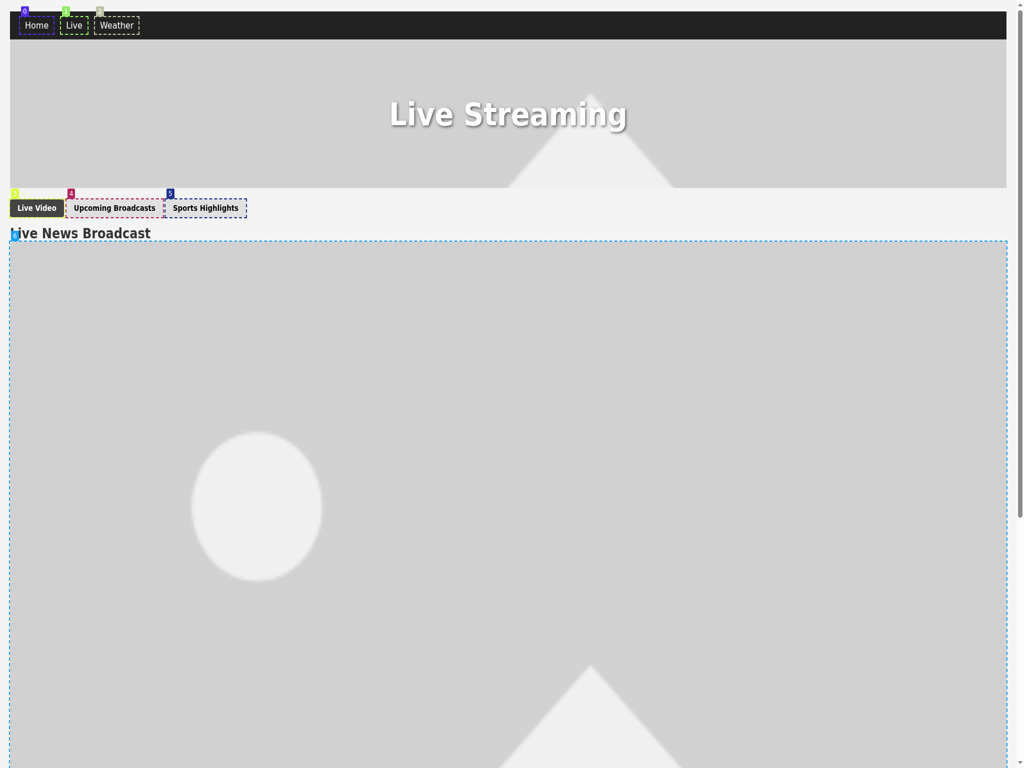} 
    & %
\textbf{Thought:} Clicking "Live Video" did not change the page content, indicating the video feed may not be functional or requires additional support from the browser. To further explore live news options, I will check "Upcoming Broadcasts" to see if there are scheduled events or programs related to news coverage.
\textbf{Action:} Click [4]  \\ \midrule
    
    4 %
    & %
      \includegraphics[width=\linewidth]{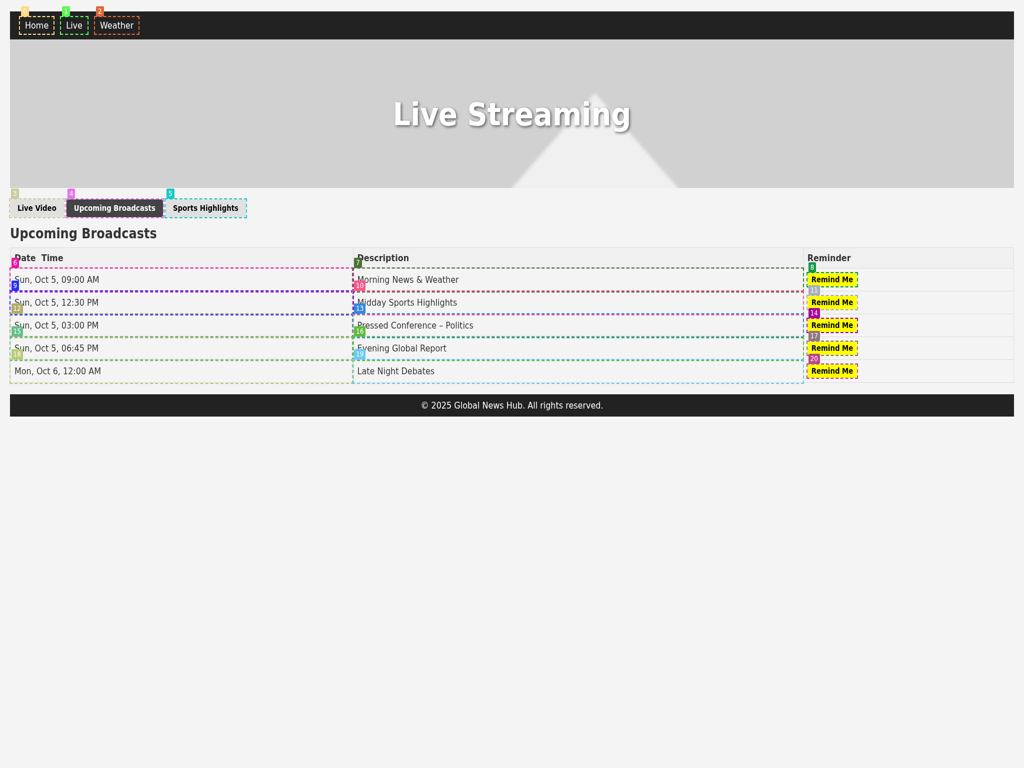} 
    & %
\textbf{Thought:} The "Upcoming Broadcasts" section provides a schedule of events, including news and weather updates, sports highlights, press conferences, global reports, and debates. Users can set reminders for these broadcasts, which is useful for keeping track of important events. To test the reminder functionality, I will set a reminder for the "Morning News \& Weather" broadcast.
\textbf{Action:} Click [8]
    \\ \midrule

    5 %
    & %
        \includegraphics[width=\linewidth]{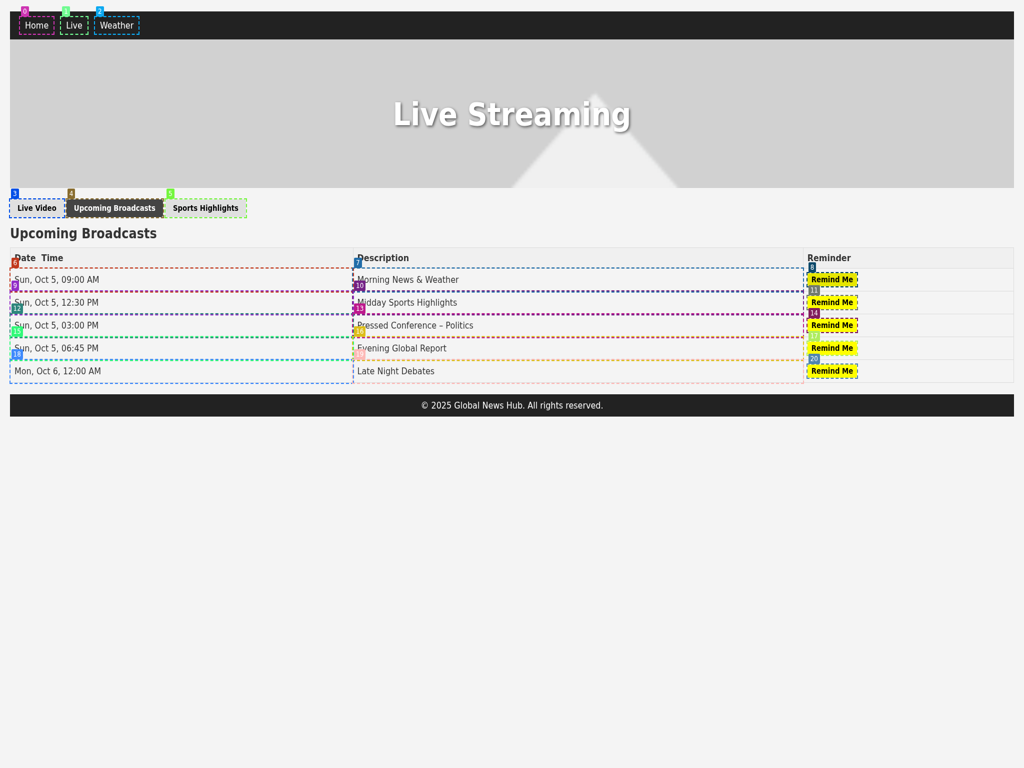} 
    & \textbf{Thought:} The pop-up message indicates that a reminder was successfully set for the "Morning News \& Weather" broadcast. This feature enhances user engagement by allowing them to keep track of specific events. To further explore the website's offerings, I will check the "Sports Highlights" section to see if it provides specific content related to sports news.
\textbf{Action:} Click [5]
    \\ \midrule

6 & \includegraphics[width=\linewidth]{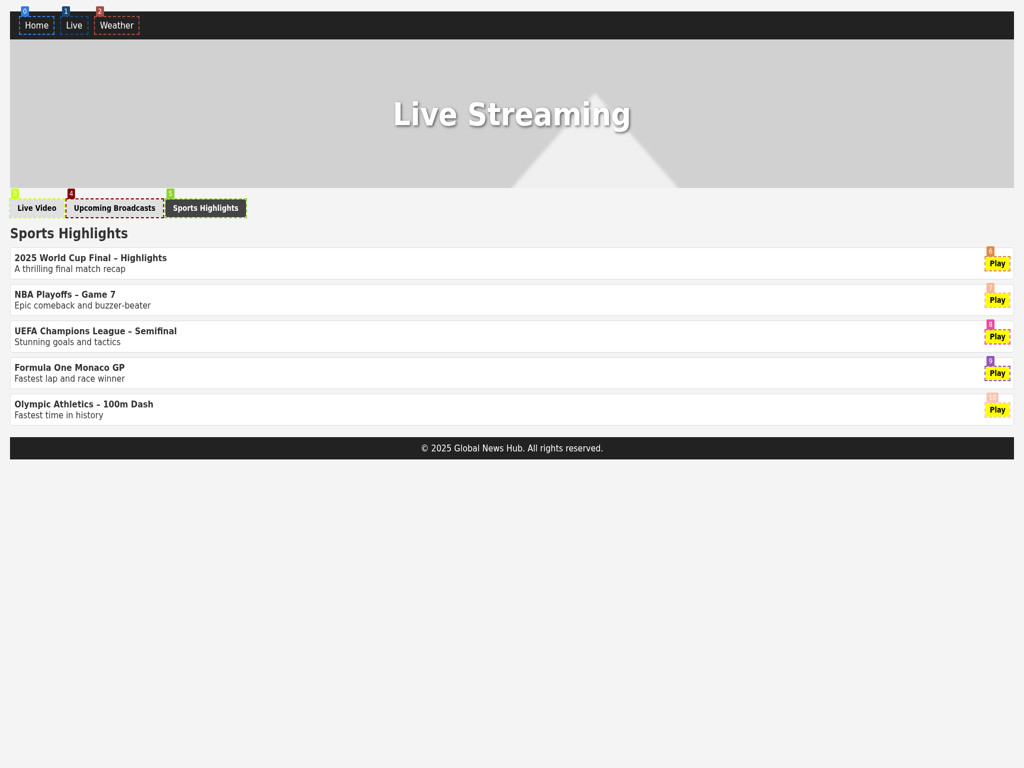} & \textbf{Thought:} The "Sports Highlights" section offers a variety of sports event recaps, including the World Cup Final, NBA Playoffs, UEFA Champions League, Formula One, and Olympic Athletics. Each event has a "Play" button, indicating video highlights are available. To test the functionality, I will click "Play" for the "2025 World Cup Final - Highlights" to view the content.
\textbf{Action:} Click [6]
\\ \midrule
7 & \includegraphics[width=\linewidth]{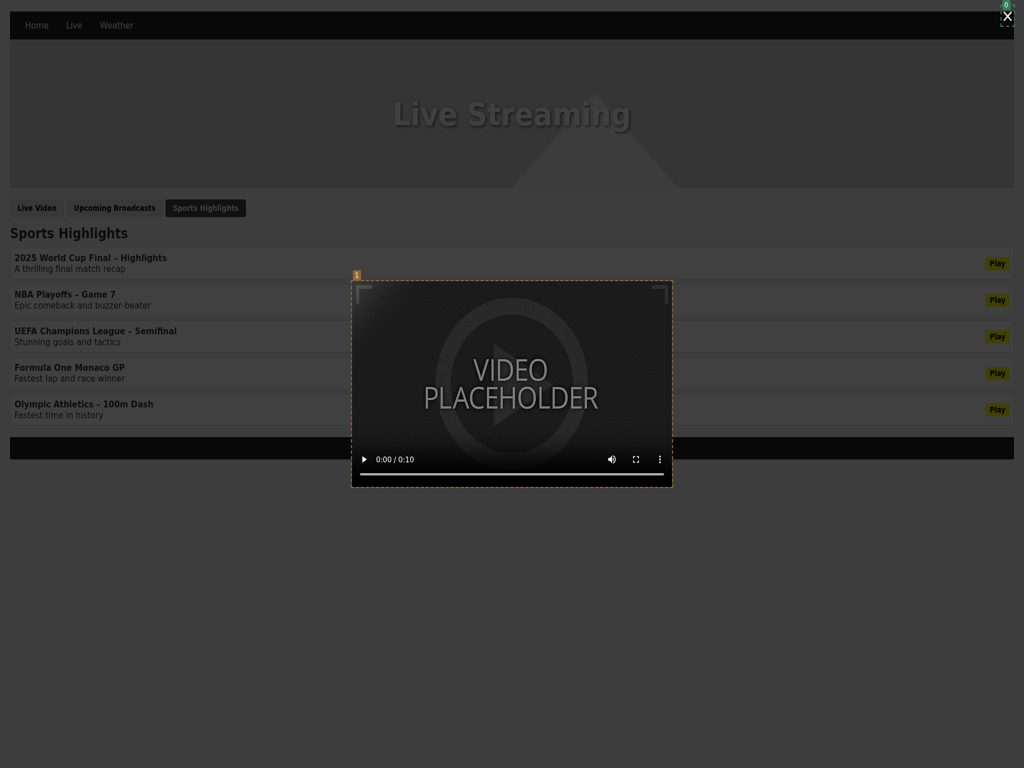} & \textbf{Thought:} Clicking "Play" for the "2025 World Cup Final - Highlights" successfully opened a video player with a placeholder video. This confirms the website's functionality for viewing sports highlights. To return to exploring other sections of the site, I will close the video player.
\textbf{Action:} Click [0] \\ \midrule
8 & \includegraphics[width=\linewidth]{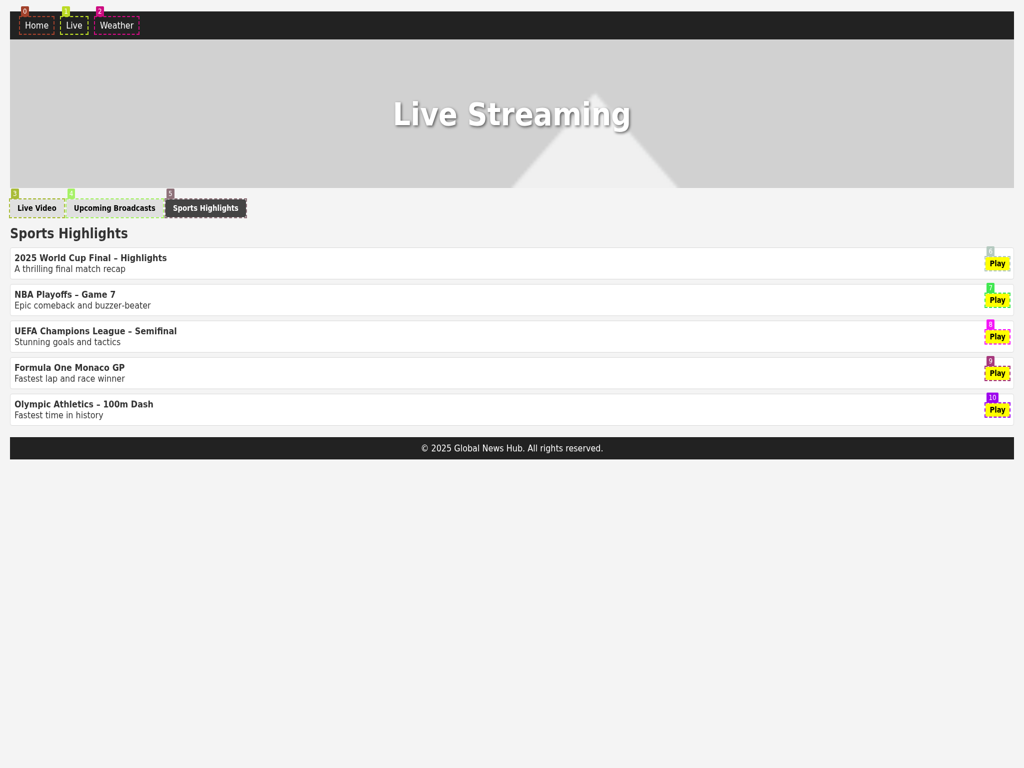} & \textbf{Thought:} The video player has been successfully closed, returning me to the "Sports Highlights" section. To explore other areas of news coverage, I will navigate to the "Weather" section to check if it provides current weather updates or forecasts.
\textbf{Action:} Click [2] \\ \midrule
9 & \includegraphics[width=\linewidth]{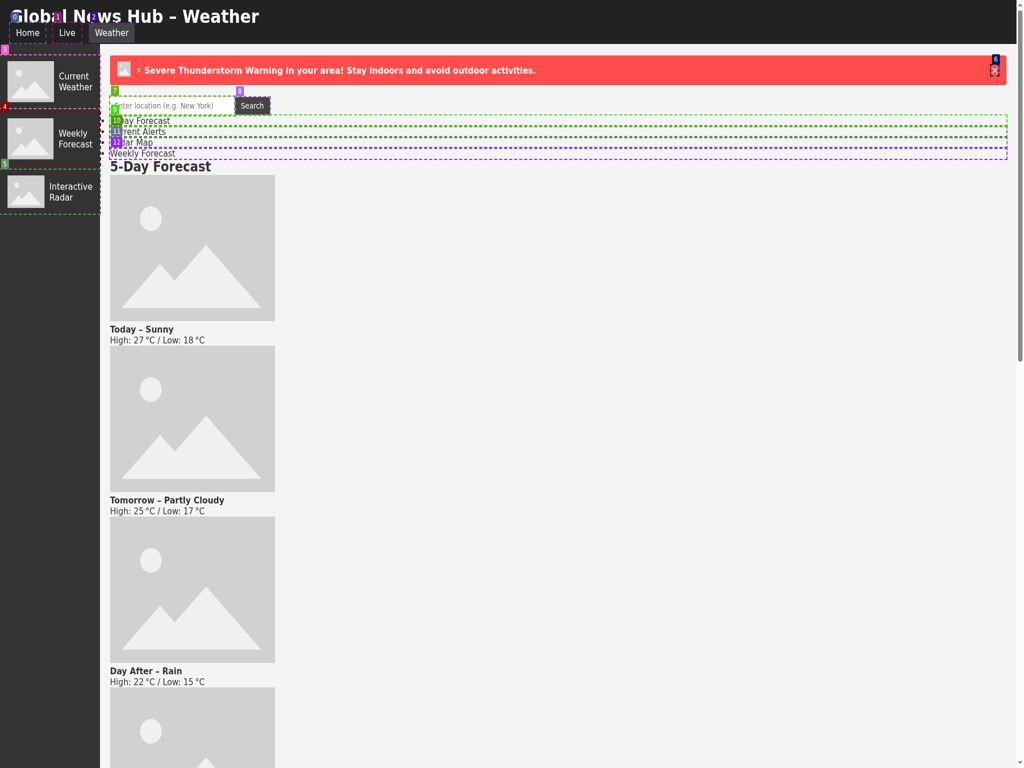} & \textbf{Thought:} The "Weather" section provides a severe thunderstorm warning, a 5-day forecast, current alerts, and an interactive radar. Users can also search for weather updates by location. To explore the current weather conditions, I will dismiss the severe weather warning message to have a clearer view of the page.
\textbf{Action:} Click [6] \\ \midrule
10 & \includegraphics[width=\linewidth]{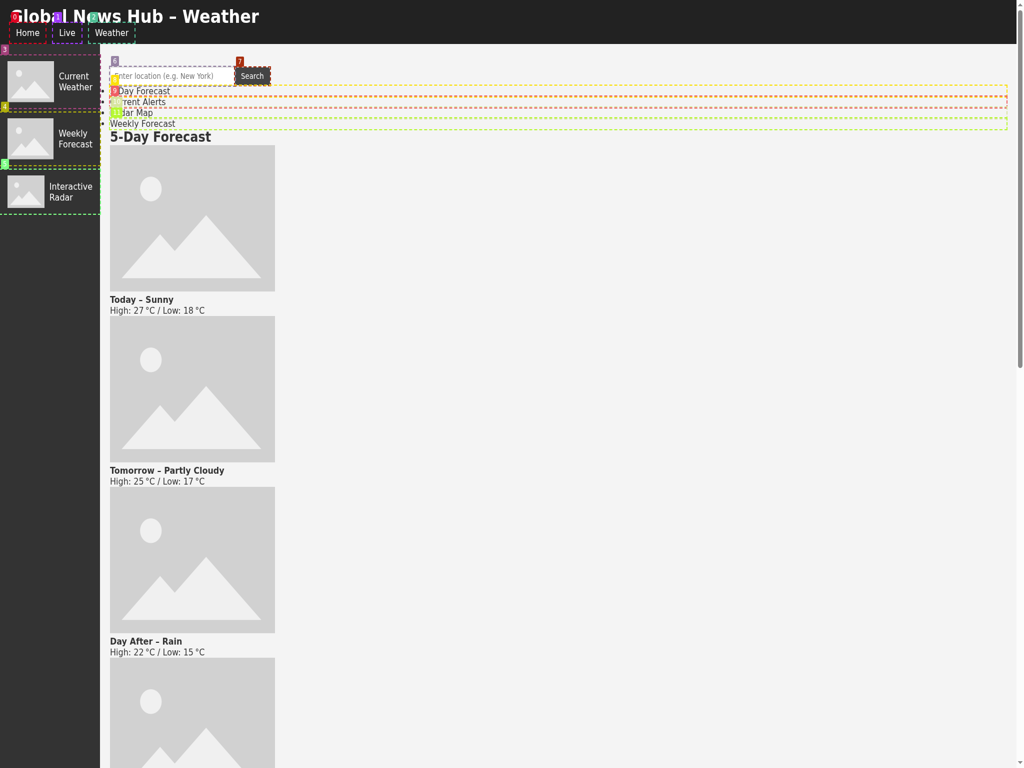} & \textbf{Thought:} The severe weather warning has been dismissed, providing a clearer view of the weather section. There's a search bar available to enter a location for specific weather updates. To test this feature, I will input a location into the search bar to see detailed weather information for that area.
\textbf{Action:} Type [6]; "New York" \\ \midrule
11 & \includegraphics[width=\linewidth]{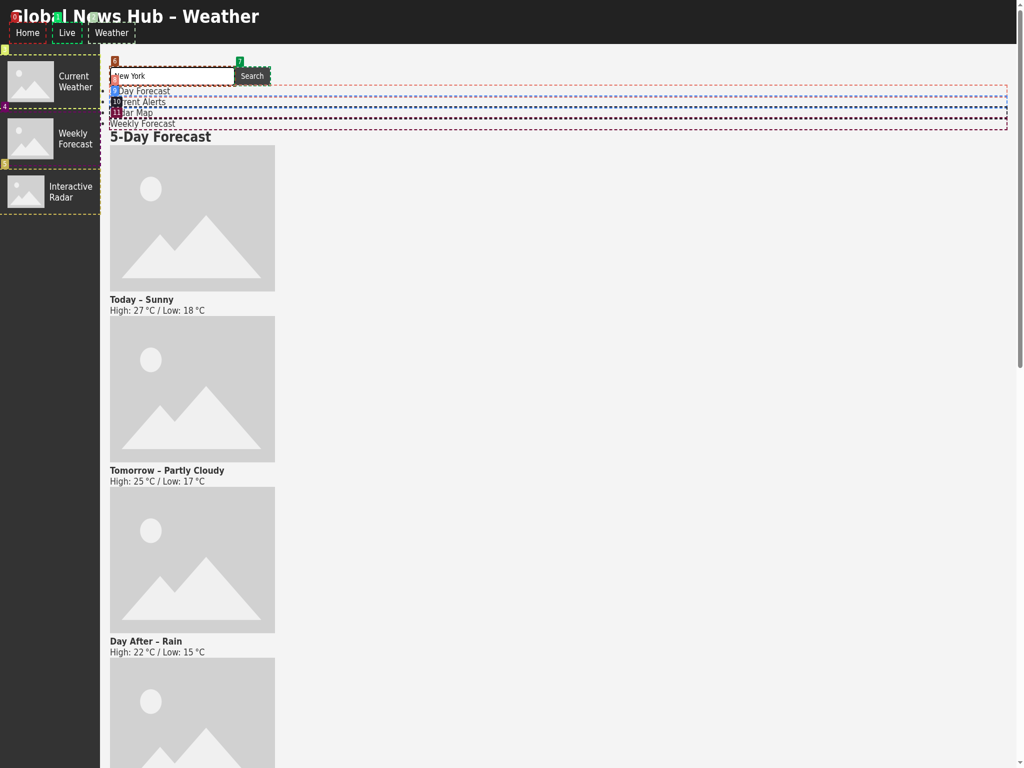} & \textbf{Thought:} I have entered "New York" into the search bar. To view the weather updates for New York, I need to click the "Search" button to apply the location filter and retrieve the relevant information.
\textbf{Action:} Click [7] \\ \midrule
12 & \includegraphics[width=\linewidth]{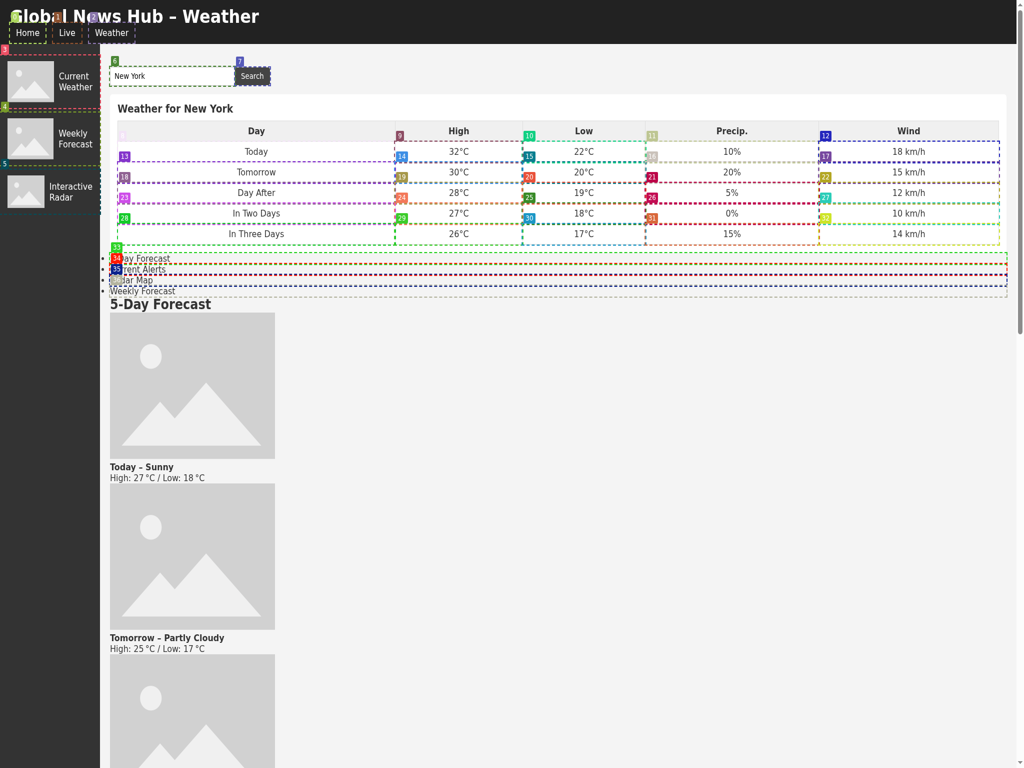} & \textbf{Thought:} After clicking the "Search" button, the weather section now displays detailed weather information for New York, including high and low temperatures, precipitation, and wind speed for the next five days. This confirms the site's ability to provide location-specific weather forecasts. To continue exploring the website's offerings, I will return to the "Home" section to see what general news coverage is available.
\textbf{Action:} Click [0] \\ \midrule
13 & \includegraphics[width=\linewidth]{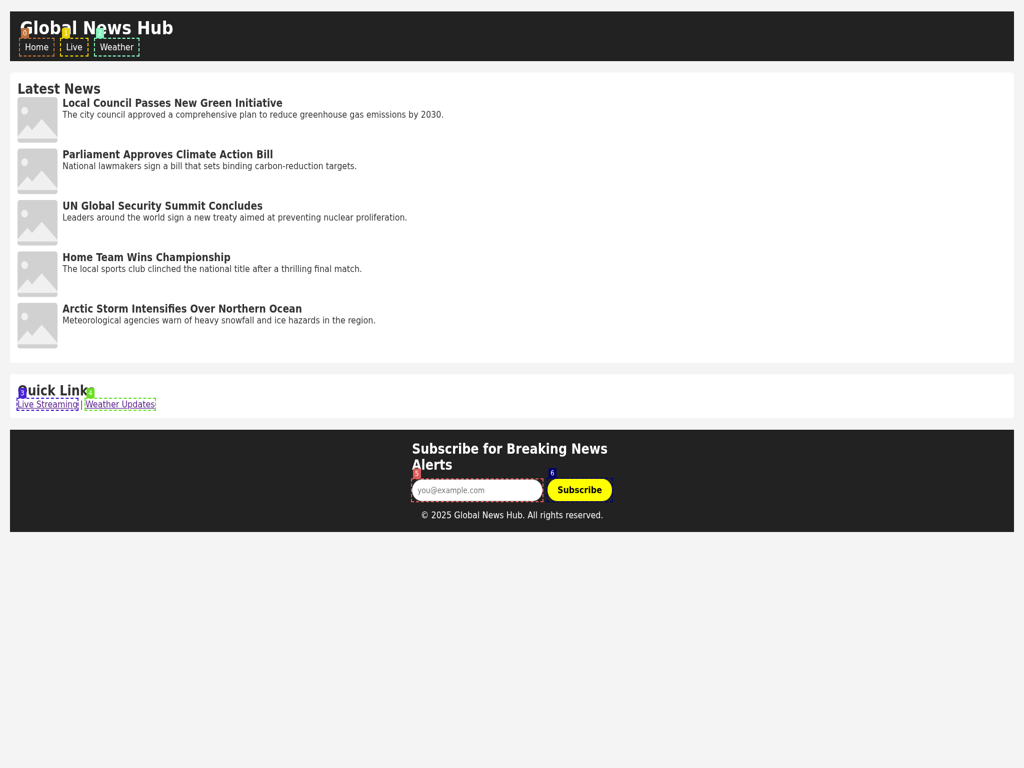} & \textbf{Thought:} Returning to the "Home" section, I see a list of the latest news articles covering various topics such as environmental initiatives, political actions, sports achievements, and weather events. There is also an option to subscribe for breaking news alerts, which is useful for users who want to stay updated. To test the subscription feature, I will enter an email address into the subscription textbox.
\textbf{Action:} Type [5]; "user@example.com" \\ \midrule
14 & \includegraphics[width=\linewidth]{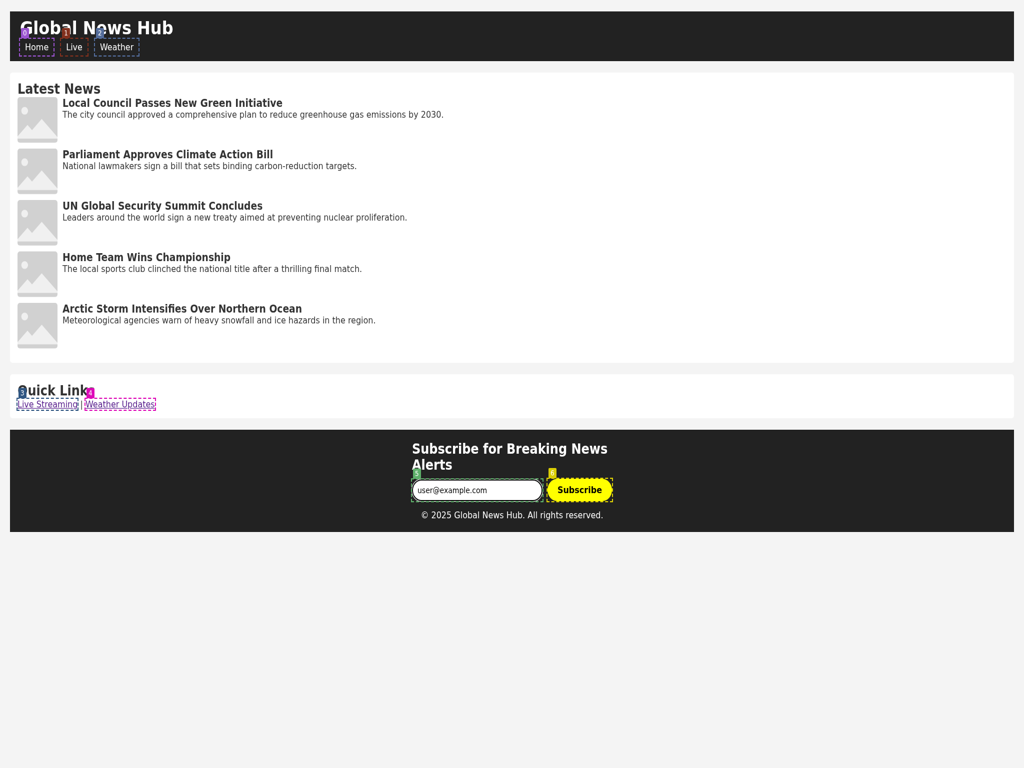} & \textbf{Thought:} I have entered the email address "user@example.com" into the subscription textbox. To complete the subscription process for breaking news alerts, I need to click the "Subscribe" button.
\textbf{Action:} Click [6] \\ \midrule
15 & \includegraphics[width=\linewidth]{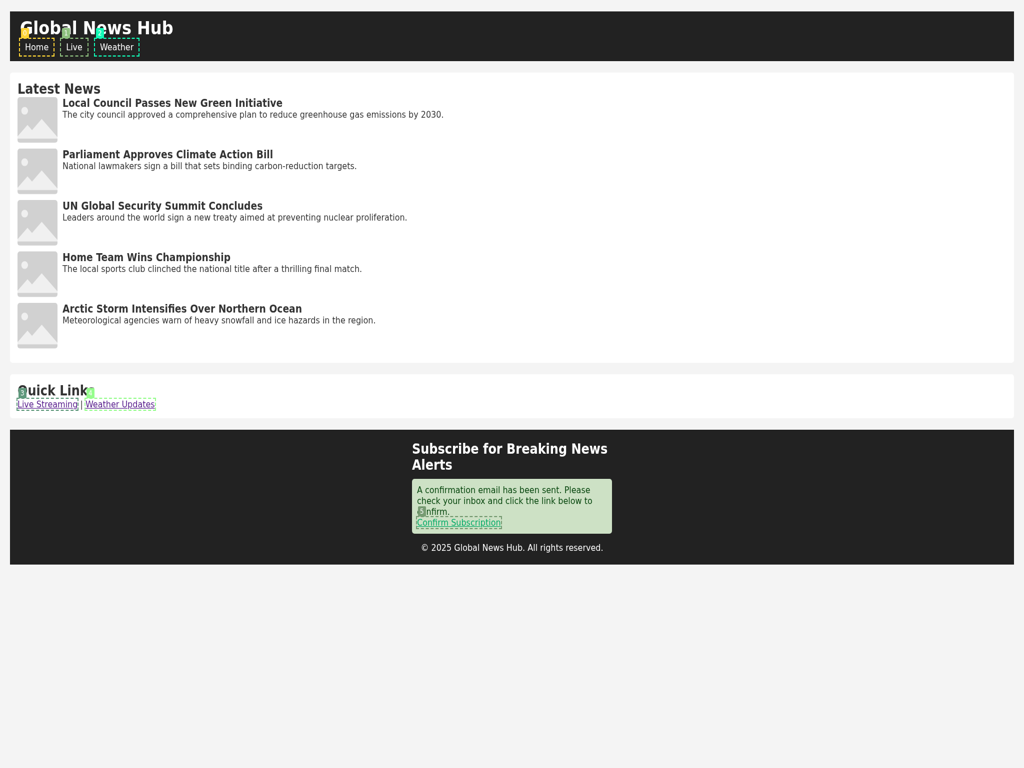} & \texttt{Step limit reached}\\
\end{longtable}

At the end of the trajectory, we prompt the model to summarize the user experience throught the trajectory. The summary for the trajectory in Table \ref{tab:my_long_table} is as follows:
\begin{lstlisting}[float=!h, basicstyle=\scriptsize\ttfamily]
The website provides a comprehensive platform for accessing news content across various domains, including local and international news, weather updates, and sports highlights. Here's a summary of my experience with the website:

1. **Navigation**: The main navigation options ("Home," "Live," "Weather") are clear and intuitive, allowing users to easily switch between different sections of the website.

2. **Live Section**: The live section offers video content and schedules for upcoming broadcasts, enhancing user engagement with live updates and reminders.

3. **Sports Highlights**: This section provides accessible video highlights for major sporting events, adding value for sports enthusiasts.

4. **Weather Section**: The weather section includes current conditions, forecasts, and alerts. It also features a location-based search function, which is practical for users seeking specific weather information.

5. **Subscription Feature**: The option to subscribe for breaking news alerts is straightforward, with clear feedback via a confirmation message upon successful subscription.

Overall, the website is user-friendly, with a clean layout and logical structure that facilitates easy access to news content. The inclusion of multimedia content and interactive features like setting reminders and searching for specific weather conditions further enhances the user experience.
\end{lstlisting}

\appsection{Prompts}

\begin{lstlisting}[caption={Prompt for evaluating \textbf{pass rate}.}, label={prompt:eval_pass_rate}]
You are an expert evaluator of built websites. You are given a test condition along with its success/failure criteria. Your task is to determine whether the website **PASS**es or **FAIL**s the test condition. You can perform the evaluation either by examining a static screenshot of the homepage or, if necessary, by navigating and interacting with the website.

At each step, I will provide you with:
1. An annotated screenshot - with **numerical labels** placed in the **top-left corner** of each web element. Each screenshot is named as screenshot_x (x=1,2,...)
2. A simplified **textual representation** of the page -- including the tag names and texts for every element annotated in the screenshot.

Your goal is to generate **ACTIONS** to perform the evaluation. If the information is sufficient to make a judgment, your ACTION should be: `ANSWER; PASS` or `ANSWER; FAIL`. Otherwise, you should generate actions to interact with the website before producing the answer. Choose one of the following valid action formats:

### Valid Actions

Action should **STRICTLY** follow the format:
- Click [Numerical_Label]
- Hover [Numerical_Label]
- Type [Numerical_Label]; [Input_Text]
- Select [Numerical_Label]; [Option_Text]
- Scroll [Numerical_Label or WINDOW]; [up or down]
- GoBack
- Upload [Numerical_Label]; [Filename]
- ANSWER; [content]

Additional actions to inspect visual details, **STRICTLY** following the format:
- Compare [Numerical_Label]; screenshot_x, screenshot_y
- ViewRaw screenshot_x
- ViewAnimation [Numerical_Label or WINDOW]; screenshot_x

### Action Guidelines

1. Execute only one action per iteration.
2. **Avoid repeating** the same action if the page does not change -- you may have chosen the wrong element or label.
3. To input text, you do **not** need to click the textbox first -- just use the `Type` action. Pressing `ENTER` is handled automatically. However, you may still need to click a search button afterward to apply a filter.
4. Clearly distinguish between textboxes and buttons -- do **not** type into a button. If no textbox is visible, consider clicking a search button to reveal it.
5. To upload a file, you do **not** need to click the upload button first -- just use the `Upload` action and specify the filename. The filename **MUST** be chosen from: `placeholder.png`, `placeholder.mp4`, `placeholder.mp3`, or `placeholder.pdf`.
6. Use `Compare` to compare the same element's visual display across two **different** screenshots. For example, compare the screenshots before (screenshot_x) and after (screenshot_y) a hover action to test hover effect.
7. Use `ViewRaw` to retrieve the high-fidelity raw screenshot without annotations.
8. Use `ViewAnimation` to view animated behavior when loading screenshot_x. To focus on a specific element, provide its numerical label.

### Evaluation Guidelines

1. You are also given the instructions for building the website, which **MAY NOT** align with the actual website structure. Use them **only as context** to help understand the website, but **always** rely on **real interactions** with the website to make the final evaluation.
2. First verify whether the feature to be tested is present on the website. If it is missing (e.g., the test condition refers to a header, but the website does not have a header), the evaluation should be FAIL.
3. If the feature to be tested is not immediately visible on the homepage but can be accessed from it, you must first navigate from the homepage to find it. **Extensively** explore the website to locate the feature, such as by scrolling down, checking menus, or following links, until you confirm whether the feature exists.
4. After navigating to the feature, if the test condition is based on **purely static** visual features such as color and layout, do not use any more actions -- directly output `ANSWER` based on the screenshot.
5. When a test involves **multiple steps or questions**, use `ANSWER` only **after** addressing all of them.
6. **Extensively** interact with the website to trigger behaviors relevant to the test condition. For example, when testing a search bar, test it with multiple inputs including both reasonable pseudo-data and generic entries like "1", "2", "a", "b".
7. The annotated screenshot is **NOT** what end users actually see. To evaluate the visual design, **always** rely on additional actions to gather visual details before making the judgment.
8. The website uses placeholder data. Do **not** judge pass/fail outcomes based on whether the displayed data reflects real-world data. 
9. The website also uses placeholder content for media (images, videos, audio, PDFs). Do **not** judge pass/fail outcomes based on the **content** of these files. However, you should evaluate their display and interactive behavior when determining pass/fail.

### Your Reply Format

Thought: {Step-by-step reasoning}
Action: {One properly formatted action}
\end{lstlisting}

\begin{lstlisting}[caption={Prompt for \textbf{simulating first-time users} in the evaluation of \textbf{usability}.}, label={prompt:eval_usability_user}]
Your task is to evaluate the **usability** of a website. You will simulate a **first-time user**: you are given only the website's high-level goal without detailed instructions. Your task is to extensively explore the website, infer what you want to accomplish as an end user, attempt the tasks, and judge how easy it is to learn and use the site to complete them.

At each step, I will provide you with:
1. An annotated screenshot - with **numerical labels** placed in the **top-left corner** of each web element.
2. A simplified **textual representation** of the page -- including the tag names and texts for every element annotated in the screenshot.

At each step, you can choose one of the following valid action formats:

### Valid Actions

Action should **STRICTLY** follow the format:
- Click [Numerical_Label]
- Hover [Numerical_Label]
- Type [Numerical_Label]; [Input_Text]
- Select [Numerical_Label]; [Option_Text]
- Scroll [Numerical_Label or WINDOW]; [up or down]
- GoBack
- Upload [Numerical_Label]; [Filename]

### Guidelines

1. Execute only one action per iteration.
2. **Avoid repeating** the same action if the page does not change -- you may have chosen the wrong element or label.
3. To input text, you do **not** need to click the textbox first -- just use the `Type` action. Pressing `ENTER` is handled automatically. However, you may still need to click a search button afterward to apply a filter.
4. Clearly distinguish between textboxes and buttons -- do **not** type into a button. If no textbox is visible, consider clicking a search button to reveal it.
5. To upload a file, you do **not** need to click the upload button first -- just use the `Upload` action and specify the filename. The filename **MUST** be chosen from: `placeholder.png`, `placeholder.mp4`, `placeholder.mp3`, or `placeholder.pdf`. 
6. The website uses placeholder for data and media (images, videos, audio, PDFs).

### Your Reply Format

Thought: {Describe image content, then perform step-by-step reasoning}
Action: {One properly formatted action}
\end{lstlisting}

\begin{lstlisting}[caption={Prompt for performing \textbf{pair-wise comparison} in the evaluation of \textbf{usability}.}, label={prompt:eval_usability_compare}]
You are an expert UX Researcher specializing in usability analysis. Your analysis must be objective, evidence-based, and grounded in established usability principles.

You will be given two user interaction trajectories, `Trajectory A` and `Trajectory B`. Your task is to perform a pairwise comparison to determine which website offers a better user experience. You will provide a detailed analysis and a final, single-word verdict for Website A: WIN, LOSE, or TIE.

### Evaluation Criteria

Base your analysis *strictly* on the following usability dimensions:

1. **Journey Success:** Did a goal emerge from the user's browsing? If so, how well did the website support their journey from discovery to potential action? Did they successfully navigate to a satisfying endpoint or did they abandon their exploration?
2. **Efficiency & Path:** How much effort (cognitive load, number of actions, backtracking) was required? Was the user's path logical and fluid, or was it disjointed and filled with unnecessary steps?
3. **Satisfaction & Friction:** How did the user feel? Identify specific moments of frustration, confusion, delight, or ease, as indicated by their thoughts.
4. **Discoverability & Clarity:** How easy was it for the user to understand the site's structure, find what's available, and learn what actions are possible? Is the value proposition clear from the start? This is especially critical for **Exploratory** tasks.

### Instructions

Follow these steps and structure your response exactly as follows.

1. **Identify Emergent Goal(s) (if any):** Briefly state the primary goal(s) the user developed during each trajectory.
2. **Analyze Trajectory A:** Summarize the user's experience on Website A, citing specific evidence for each of the four evaluation criteria.
3. **Analyze Trajectory B:** Do the same for Website B.
4. **Comparative Analysis:** Directly compare Website A and Website B on each of the criteria. State which website performed better and why.
5. **Final Verdict & Justification:** Conclude with a verdict for Website A and a concise justification highlighting the most critical factors. Formulate your final verdict as follows:

VERDICT: [WIN, LOSE, or TIE]
\end{lstlisting}

\begin{lstlisting}[caption={Prompt for \textbf{textual instruction simulator} for functionality-type user intents. The {\color{darkgold}gold} text refers to user intent, and the {\color{LinkBlue}blue} text refers to the current code state.}, label={prompt:textual_user_f}]
You are provided with instructions to **add or refine functionalities** of a website. Your task is to refine the instructions so they are accurate, specific, and actionable. Use the code of the existing website as context.
1. **Verify component or functionality references.** These references typically involve **where** to implement the new feature, or **how** to navigate to the new feature from the homepage.
   * If a referenced component/functionality exists, refine vague references to be more precise.
     * Example: Change "add a button to the main menu" to "add a button to the dark gray navigation bar at the top".
   * If it does **not** exist, clearly note that it must be implemented before the instruction can be applied.
2. **Clarify vague instructions.** If an instruction is not specific enough, make it more concrete based on the actual website.
3. **Avoid code-level details.** Do not reference class names, HTML attributes, color hex codes, or other implementation-specific identifiers.

Your refined instructions should be a short paragraph (3-6 sentences). Start your refined instructions with
**Response:**

# Instructions to Refine

(*@\textcolor{darkgold}{USER INTENT}@*)

# Code for Existing Website

(*@\textcolor{LinkBlue}{\#\# filename1}@*)

(*@\textcolor{LinkBlue}{file content 1}@*)

(*@\textcolor{LinkBlue}{\#\# ...}@*)
\end{lstlisting}

\begin{lstlisting}[caption={Prompt for \textbf{textual instruction simulator} for design-type user intents. The {\color{darkgold}gold} text refers to user intent, and the {\color{LinkBlue}blue} text refers to the current code state.}, label={prompt:textual_user_d}]
You are provided with instructions to **refine the visual design** of a website. Your task is to refine the instructions so they are accurate, specific, and actionable. Use the code of the existing website as context.
1. **Check for already implemented instructions.** Compare each instruction to the current state of the website. If an instruction is already implemented, remove it from the refined list.
   * Example: If the instruction says "Use a two-column layout" but the site already has a two-column layout, remove that instruction.
   * Keep all unfulfilled instructions exactly as they are - **do not omit or change them** unless covered by steps 2-3 below.
2. **Verify component or functionality references.**
   * If a referenced component/functionality exists, refine vague references to be more precise.
     * Example: Change "the button" to "the green submit button in the bottom left".
   * If it does **not** exist, clearly note that it must be implemented before the instruction can be applied.
3. **Clarify vague instructions.** If an instruction is not specific enough, make it more concrete based on the actual website.
4. **Avoid code-level details.** Do not reference class names, HTML attributes, color hex codes, or other implementation-specific identifiers.

Your refined instructions should be a short paragraph (~5 sentences). Start your refined instructions with
**Response:**

# Instructions to Refine

(*@\textcolor{darkgold}{USER INTENT}@*)

# Code for Existing Website

(*@\textcolor{LinkBlue}{\#\# filename1}@*)

(*@\textcolor{LinkBlue}{file content 1}@*)

(*@\textcolor{LinkBlue}{\#\# ...}@*)
\end{lstlisting}

\begin{lstlisting}[caption={Prompt for \textbf{visual instruction simulator}. The {\color{darkgold}gold} text refers to user intent. The screenshots for every rendered page along with their filenames (in the format of \texttt{screenshot\_name.png}) are appended in the following messages.}, label={prompt:visual_user}]
You are an expert drawing agent with a series of drawing tools based on `matplotlib`. Your task is to visualize a series of user instructions about implementing or refining the {{functionality|visual design}} of a website.

For each page in the website, you are given:
* A screenshot (with its filename),
* The HTML code, and
* The coordinates of elements in the page.

You can load and display a screenshot with:
```python
from PIL import Image
ax.imshow(Image.open("screenshot_name.png"))
```

# User Instructions

(*@\textcolor{darkgold}{USER INTENT}@*)

# Tool Documentation

You have the following four tools:
1. Subplot organization
2. Layout visualization
3. Shape drawing
4. Text annotation

You'll generate python code to call these tools - I'll explain the tools one by one. Your code should be wrapped within ``python ```. Make sure to import necessary tools and libraries you want to use.

## Subplot Organization

When visualizing the instructions, you may need to work with multiple subplots:
1. **First subplot - always the screenshot of relevant page**. Use shapes and text annotations to link each instruction to its corresponding component. For example:
   * If the instruction states, "Use a dark background color with white text for the menu items", add annotations to the menu bar.
   * If it states, "Display the discount price in a bright color", annotate the product card. 
   * If it states, "Use a circular profile photo with a thin border", annotate the profile picture accordingly.
2. **Revised layout subplot(s)** - for layout changes, if necessary.
   1. In the first subplot, highlight the components to be moved and indicate their intended movement.
   2. In the later subplot(s), show the components in their **new positions** with shapes and text annotations - using the **same annotation color** for each component in both the original screenshot and the updated layout views for clarity.
   3. To show the updated layout: (1) If the UI changes are minor, reuse the screenshot and mark it with shapes showing the new positions. (2) For major changes, use the HTML layout visualization tool to create a simplified updated layout.
3. **State transition subplot(s)** - for interaction-driven changes, if necessary
   1. In the first subplot, highlight the interactive component(s) that trigger the transition and annotate interaction type.
   2. In the new subplot, show the resulting end state and clearly indicate what changed.
4. **Titles**: Each subplot should have a **clear and descriptive title** to clearly indicate the relationship and transitioning between subplots (e.g., "Current Layout", "Proposed Layout", "After Clicking `Submit`").
5. **Multiple pages**: If the instructions involve multiple pages, you can allocate one or multiple subplots for each page following the instructions above. Annotations for each page will be based on its own screenshot.

You may create subplots **only in your first drawing turn**. Before starting, decide exactly how many subplots you need and what **specific page state** each will represent. Once created, each subplot is fixed to a single, distinct view - you cannot add or remove subplots later.

Use matplotlib to create multiple subplots, and use each element of `axes` for further drawing:
```python
fig, axes = plt.subplots(n_row, n_col)
```

Note: For each subplot, **NEVER** directly draw on top of blank canvas background! You **must** define a background using one of the following methods:
1. Screenshot background: If you are visualizing the existing page or interaction flow, use the screenshot: `ax.imshow(Image.open("current_screenshot.png"))`
2. HTML layout background: If you are visualizing a new layout or interface, use the layout visualization tool: `layout_visualization(html_code, ax)`. After this call, wait for coordinate data before adding shapes or annotations.

## Layout Visualization

To visualize the components required in the instructions, you should create a **minimal** HTML layout **only use boxes and text**. Avoid colors, fonts, and detailed styling.

**Important**: You may receive reference HTML in the input. These are for context only - **do not** copy or reuse them in your `layout_visualization` call.

Use the following to visualize HTML layout in a given `ax`:
```python
html_code = """<HTML code here>"""
layout_visualization(html_code, ax)
```
Assume the function `layout_visualization` is already implemented - do not redefine it. After you call it, I will provide coordinates for each element in the next round. You can then use those to draw shapes or text annotations in future steps. **Do not** perform any annotations on `ax` before receiving the coordinates.

## Shape Drawing

Use `matplotlib.patches` to draw rectangles, circles, ellipses, or polygons. Shapes are typically used for: (1) Highlighting specific regions for which you have instructions; or (2) Indicating the addition of new components, such as a circle for an icon or a rectangle for a new block.

Examples:
```python
new_shape = patches.Rectangle((x, y), width, height, facecolor='none', edgecolor='#??????')  # draw rectangle
new_shape = patches.Circle((x_center, y_center), radius, facecolor='none', edgecolor='#??????')  # draw circle
new_shape = patches.Ellipse((x_center, y_center), width, height, facecolor='none', edgecolor='#??????')  # draw ellipse
new_shape = patches.Polygon([[x1, y1], [x2, y2], [x3, y3]], closed=True, facecolor='none', edgecolor='#??????')  # draw polygon
```

and then draw the shape:
```python
ax.add_patch(new_shape)
```

Shape drawing is usually paired with text annotation - you first draw a shape to highlight a specific component, then annotate it with text. When doing so, follow these color guidelines for clarity and visual coherence:
1. **Consistent color for pairing**: Use the same `edgecolor` for the shape and `color` for the text annotation to visually link them as a pair 
2. **Shared annotation across components**: If a single annotation (e.g., "add shadow effect") applies to multiple components, avoid repeating the same annotation for each one. Instead, draw shapes around **all relevant components** using the same distinctive color, and annotate **just one** of them with a note explaining that the annotation applies to all shapes of that color. 
3. **Distinct colors for distinct meanings**: If you are applying **multiple different annotations**, use **different colors** for each group of shape/text pairs to clearly distinguish between them.

## Text Annotation

To annotate text, **NEVER use `ax.text` or `ax.annotate` directly**! **Always** use the following function:
```python
text_annotation(ax, text, new_shape, color='#??????')
```
* `new_shape` should be a Rectangle, Circle, Ellipse, or Polygon (as described in the Shape Drawing section) you want to annotate.
* This function draws an arrow pointing to the shape, with a labeled box at the arrow's origin. The box and arrow will both use the specified `color`, creating a visually consistent annotation.
* Placement and overlap are automatically managed to ensure clarity and avoid visual clutter.
* The `text_annotation` function is already implemented - do not redefine it.

## Multi-Turn Tool Calling

You can complete the drawing in a single turn or across multiple turns. After each turn, I will provide the current figure. If you've called `layout_visualization` in the turn, I will also provide the coordinates for every element in the HTML - so if necessary, you can draw annotations over the HTML visualization.

Most tasks can be completed in one turn. However, if you need to annotate elements based on coordinates of HTML layout, it is acceptable to use multiple turns.
\end{lstlisting}

\end{document}